%% file: CVPR_camera.tex
\documentclass[10pt,twocolumn,letterpaper]{article}

\usepackage{cvpr}
\usepackage{times}
\usepackage{epsfig}
\usepackage{graphicx}
\usepackage{amsmath}
\usepackage{amssymb}
\usepackage{mathrsfs}
\usepackage{algorithm}
\usepackage{algpseudocode}
\usepackage{multirow}
\usepackage{booktabs}
\usepackage{verbatim}
\usepackage{color}  
\usepackage{enumitem}

\usepackage[breaklinks=true,bookmarks=false]{hyperref}

\cvprfinalcopy 

\def\cvprPaperID{****} 

\ifcvprfinal\fi

\newlength\savewidth
	
\newenvironment{itemize*}%
{\begin{itemize}%
		\setlength{\itemsep}{0pt}%
		\setlength{\parskip}{5pt}}%
{\end{itemize}}

\makeatletter\renewcommand\paragraph{\@startsection{paragraph}{4}{\z@}
	{.25em \@plus1ex \@minus.2ex}{-.5em}{\normalfont\normalsize\bfseries}}\makeatother

\newcommand{\degree}{^\circ}

\makeatletter
\newcommand{\thickhline}{%
	\noalign {\ifnum 0=`}\fi \hrule height 1pt
	\futurelet \reserved@a \@xhline
}
\makeatother

\begin{document}

\title{Universal Physical Camouflage Attacks on Object Detectors}

\author{
Lifeng Huang\textsuperscript{1,2} Chengying Gao\textsuperscript{1} Yuyin Zhou\textsuperscript{3} Cihang Xie\textsuperscript{3} Alan Yuille\textsuperscript{3} Changqing Zou\textsuperscript{4,5} Ning Liu\textsuperscript{1,2~\thanks{Corresponding author.}}
\\
\textsuperscript{1}School of Data and Computer Science, Sun Yat-sen University\\
\textsuperscript{2}Guangdong Key Laboratory of Information Security Technology\\
\textsuperscript{3}Department of Computer Science, The Johns Hopkins University\\
\textsuperscript{4}Max Planck Institute for Informatics,
\textsuperscript{5}University of Maryland, College Park\\
\tt\small huanglf6@mail2.sysu.edu.cn,
\{mcsgcy,~liuning2\}@mail.sysu.edu.cn, \\
\tt\small\{zhouyuyiner,~cihangxie306,~alan.l.yuille,~aaronzou1125\}@gmail.com
}

\maketitle
\thispagestyle{empty}

\input{0_abs.tex}

\input{1_intro.tex}

\input{2_related.tex}

\input{3_method.tex}

\input{4_evaluation.tex}

\input{5_discussion.tex}

\input{6_conclusion.tex}

{\small
\bibliographystyle{ieee_fullname}
\bibliography{egbib}
}

\input{supp.tex}

\end{document}

%% file: 0_abs.tex
\begin{abstract}
	In this paper, we study physical adversarial attacks on object detectors in the wild. Previous works mostly craft instance-dependent perturbations only for rigid or planar objects.
	To this end, we propose to learn an adversarial pattern to effectively attack \textbf{all instances belonging to the same object category}, referred to as Universal Physical Camouflage Attack (UPC).
	Concretely, UPC crafts camouflage by jointly fooling the region proposal network, 
	as well as misleading the classifier and the regressor to output errors. In order to make UPC effective for non-rigid or non-planar objects, we introduce a set of transformations for mimicking deformable properties.
	We additionally impose optimization constraint to make generated patterns look natural to human observers.
	To fairly evaluate the effectiveness of different physical-world attacks,
	we present the first standardized virtual database, \textbf{\emph{AttackScenes}}, which simulates the real 3D world in a controllable and reproducible environment.
	Extensive experiments suggest the superiority of our proposed UPC compared with existing physical adversarial attackers not only in virtual environments (\textbf{\emph{AttackScenes}}), but also in real-world physical environments. 
	Code and dataset are available at \url{https://mesunhlf.github.io/index_physical.html}.

\end{abstract}

%% file: 1_intro.tex
\begin{figure}[t]
\begin{center}
\includegraphics[width=0.9\linewidth]{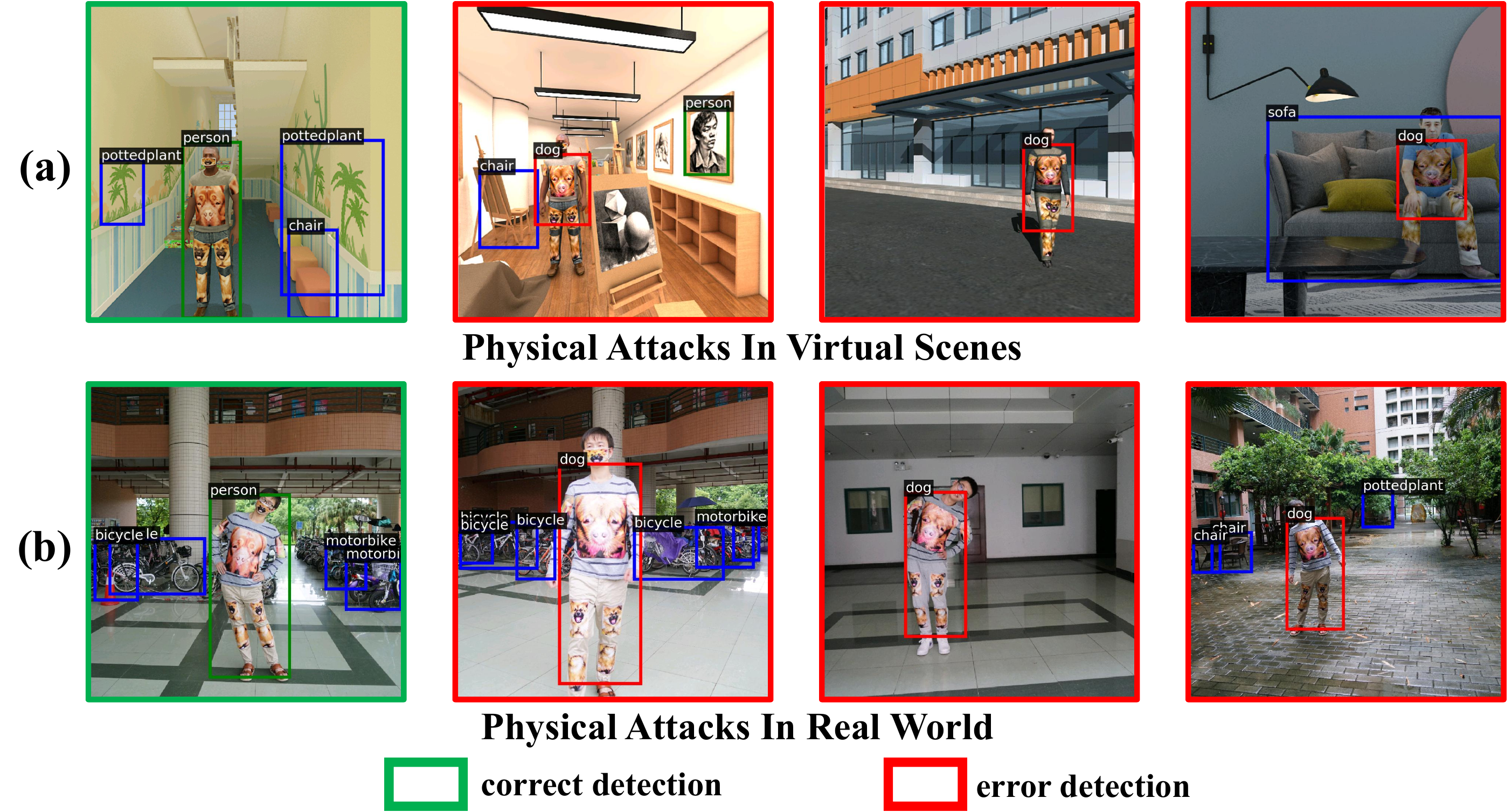}
\end{center}
\vspace{-1em}
   \caption{
   \textbf{Fooling the object detector, faster r-cnn, in the physical space.} 
   (a) Physical attacks (UPC) in virtual scenes and (b) Physical attacks (UPC) in real world. Column 1 shows detection results with natural patterns. Column 2-4 display results with camouflage patterns under different viewing conditions.
   }
\label{fig:example}
\vspace{-1.5em}
\end{figure}

\section{Introduction}
Deep neural networks (DNNs) have achieved outstanding performances on many computer vision tasks~\cite{szegedy2016rethinking,he2016deep,huang2017densely}.
Nonetheless, DNNs have been demonstrated to be vulnerable to adversarial examples \cite{szegedy2013intriguing} --- maliciously crafted inputs that mislead DNNs to make incorrect predictions,
which present potential threats for the deployment of DNN-based systems in the real world.

Adversarial attacks~\cite{moosavi2016deepfool,bhagoji2018practical} in general can be divided into the following categories:
1) \textbf{digital attacks}, which mislead DNNs by modifying the input data directly in the digital space
(\eg, pixel value~\cite{moosavi2016deepfool,ilyas2018black,liu2018dpatch}, text content~\cite{li2018textbugger,samanta2017towards});
2) \textbf{physical attacks}, which attack DNNs by altering \textbf{visible} characteristics of an object (\eg, color~\cite{sharif2016accessorize}, appearance~\cite{evtimov2017robust}) in the physical world.
Current mainstream works focus on the digital domain, which can be hardly transferred to the real world due to the lack of considering physical constraints (\eg, invariant to different environmental conditions such as viewpoint, lighting)~\cite{evtimov2017robust}. 
In this paper, we study adversarial attacks in the physical world, which are more threatening to real-world systems~\cite{kurakin2016adversarial}. Compared with previous works~\cite{brown2017adversarial,jan2019connecting,athalye2017synthesizing} which mostly focus on attacking image classification systems, we consider the far more realistic computer vision scenario, \ie, object detection.

Though prior works have revealed the vulnerability of object detectors to adversarial perturbations in the real world~\cite{chen2018shapeshifter,song2018physical,zhang2018camou}, there are several limitations: 
(1) focusing on only attacking a specific object (\eg a stop sign~\cite{evtimov2017robust,chen2018shapeshifter}, commercial logo~\cite{sitawarin2018darts} or car~\cite{zhang2018camou});
(2) generating perturbations only for rigid or planar objects (\eg, traffic sign, vehicle body, board~\cite{thys2019fooling}), 
which can be less effective for complex objects (articulated non-rigid or non-planar objects, \eg, human). 
(3) constructing meaningless which lack semantics and appear unnatural for human observers (\ie,  noisy or mosaic-like texture)~\cite{chen2018shapeshifter,thys2019fooling,zhang2018camou}; and 
(4) a unified evaluation environment is missing, which makes it difficult to make fair comparisons between different attacks.

To address these issues, we present Universal Physical Camouflage Attack (UPC), which
constructs a universal camouflage pattern to hide objects from being detected or to misdetect objects as the target label. 
Unlike former works which generate instance-level perturbations, UPC constructs a universal pattern to attack all instances that belong to the same category (\eg, person, cars) via jointly attacking the region proposal network, the classifier and the regressor. To efficiently handle the deformations of complex objects in the physical world, we propose to model their deformable characteristics as well as external physical environments in UPC.
Specifically, the internal properties
are simulated by applying various geometric transformations (\eg, cropping, resizing, affine homography). 
We impose additional optimization constraint to encourage the visual resemblance between generated patterns and natural images, which we refer to as \emph{semantic constraint}.
As shown in Fig.~\ref{fig:example}, these camouflage patterns are visually similar to natural images and thus can be regarded as texture patterns on object surfaces such as human accessories/car paintings. 
The overall pipeline is illustrated in Fig.~\ref{fig:pipeline}.

To fairly evaluate the effectiveness of different physical attacks, we provide the first standardized synthetic dataset, \ie, \emph{\textbf{AttackScenes}}. All experimental data is generated under strict parametric-controlled physical conditions to ensure that the evaluation is reliable under virtual settings. 

The contributions of our work are four-fold:
\begin{itemize*}
\item 
UPC constructs a universal camouflage pattern for effectively attacking object detectors based on the fact that the generated pattern can be naturally camouflaged as texture patterns on object surfaces such as human accessories/car paintings.

\item 
We present the first standardized dataset, \emph{\textbf{AttackScenes}}, which is simulates the real 3D world under controllable and reproducible settings, to ensure that all experiments are conducted under fair comparisons for future research in this domain.
\item To make UPC effective for articulated non-rigid or non-planar objects, we introduce additional transformations for the camouflage patterns to simulate their internal deformations.
\item Our proposed UPC not only achieves state-of-the-art result for attacking object detectors in the wild, but also exhibits well generalization and transferability among different models.
\end{itemize*}

%% file: 2_related.tex
\section{Related Works}
\paragraph{Universal Adversarial Attack.}
\emph{Image-agnostic} attack, \ie, universal adversarial attack~\cite{moosavi2017universal, khrulkov2018art}, is defined as an attack which is able to fool different images with a single global pattern in the digital domain.  
Here we extend this definition to the physical domain and define \emph{instance-agnostic} perturbations as universal physical attacks for object detectors. Unlike former physical attack methodologies which craft instance-level patterns, our goal is to generate a single camouflage pattern to effectively attack \emph{all instances of the same object category} given different physical scenes.

\paragraph{Physical Attacks.}
Stem from the recent observation that printed adversarial examples can fool image classifiers in the physical world~\cite{kurakin2016adversarial,jan2019connecting}, efforts
have been investigated to study how to construct ``robust'' adversarial examples in the real physical world.
For instance, Athalye~\etal~\cite{athalye2017synthesizing} propose to construct 3D adversarial objects by attacking an ensemble of different image transformations;
Sharif~\etal~\cite{sharif2016accessorize} successfully attack facial recognition systems by printing textures on eyeglasses;
Evtimov~\etal~\cite{evtimov2017robust} use poster, sticker and graffiti as perturbations to attack stop signs in the physical world.
Zeng \etal \cite{zeng2017adversarial} apply computer graphics rendering methods to perform attacks in the 3D physical world.
In addition, adversarial attacks also extend to fool tracking system and Re-Identification models~\cite{Wang_2019_ICCV,Wiyatno_2019_ICCV}.

\begin{figure*}[t!]
\centering
\includegraphics[width=0.95\textwidth]{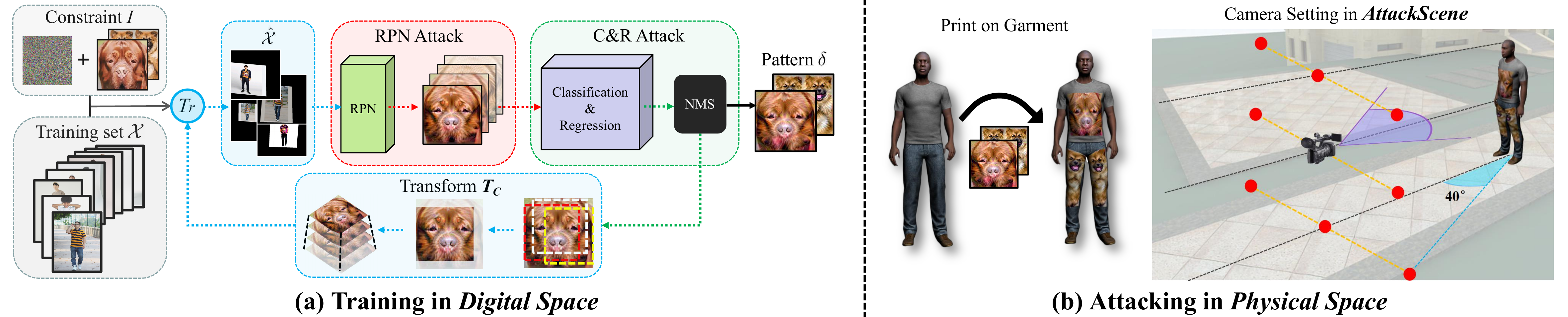}
\caption{The overall pipeline of UPC. (a) training the camouflage patterns in digital space; (b) attacking the target in physical space.}
\vspace{-1.2em}
\label{fig:pipeline}
\end{figure*}

Recently, physical attacks have also been studied for the more challenging scenario of object detection.
Song \etal~\cite{song2018physical} propose a disappearance and creation attack to fool Yolov2 \cite{redmon2017yolo9000} in traffic scenes.
Chen \etal~\cite{chen2018shapeshifter} adopt the expectation over transformation method \cite{athalye2017synthesizing} to create more robust adversarial stop signs, which mislead faster r-cnn~\cite{ren2015faster} to output errors. 
Zhang \etal~\cite{zhang2018camou} learn the clone network to approximate detectors under black-box scenerio.
However they cannot be effectively applied to non-rigid or non-planar objects since they only focus on simulating external environment conditions, \eg, distances or viewpoints, for attacking object detectors. In addition, these approaches generate instance-dependent patterns which exhibit less semantics and therefore the perturbed images are usually unnatural and noisy. Different from these works, our method constructs a universal semantic pattern which makes the perturbed images visually similar to natural images. Meanwhile, we introduce additional transformations to simulate the deformable properties of articulated non-rigid or non-planar objects.
A detailed comparison with former methods is summarized in Table.~\ref{table:method_compare}.

\newcommand{\personalsize}{\fontsize{6.2pt}{8.5pt}\selectfont}
\begin{table}[t]
	\personalsize
	\centering
	\caption{Comparison with existing methods.}
    \begin{tabular}{ccccccc}
	\toprule
	Methods & Rigid & Non-Rigid & Planar & Non-Planar & Universal & Semantic\\ 	
	\midrule
	\cite{chen2018shapeshifter} & \checkmark & & \checkmark & & \\
	\cite{song2018physical} & \checkmark & & \checkmark & & \\
	\cite{zhang2018camou} & \checkmark & & \checkmark & \checkmark & \\
	Ours & \checkmark &  \checkmark & \checkmark & \checkmark & \checkmark & \checkmark \\
	
	\bottomrule
	\vspace{-4em}
\end{tabular}
\label{table:method_compare}
\end{table}

%% file: 3_method.tex
\section{Methodology}
\subsection{Overview}
Our goal is to attack object detectors by either hiding the object from being detected, or fooling detectors to output the targeted label. 
Without loss of generality, we use ``person" category as an example to illustrate our method.

\paragraph{Training framework of UPC in Digital Space.}~We attack faster-rcnn \cite{ren2015faster}, a two-stage detector, under white-box settings. In the first stage, the region proposal network is employed to generate object proposals. In the second stage, the detector selects top-scored proposals to predict labels.
We propose to craft a universal pattern for faster-rcnn by jointly fooling the region proposal network to generate low-quality proposals, \ie, reduce the number of valid proposals, as well as misleading the classifier and the regressor to output errors. Simply misleading predictions of the classification head cannot produce satisfying results (discussed in Sec.~\ref{sec:virtual}) because it can be impractical to attack enormous candidate proposals simultaneously. Extensive experimental results also validate that the joint attack paradigm demonstrates stronger attacking strength than simply attacking the classifier as in prior methods~\cite{chen2018shapeshifter,evtimov2017robust} (Table~\ref{table:compare_table}). Furthermore, to deal with complex objects, we propose to simultaneously model both internal deformable properties of complex objects and external physical environments. The internal attributes of objects, \ie, deformations, are simulated by a series of geometric transformations.
As illustrated in Fig.~\ref{fig:pipeline}(a), UPC consists of 3 steps:
\begin{itemize}[leftmargin=*]
    \setlength\itemsep{0.1em}
	\item \textbf{Step 1}. A set of perturbed images are synthesized by simulating external physical conditions (\eg, viewpoint) as well as internal deformations of complex objects. An additional optimization constraint is imposed to make the generated patterns semantically meaningful (Sec.~\ref{section3.4}). 
	
	\item \textbf{Step 2}. Initial adversarial patterns are generated by attacking the RPN, which results in a significant drop of high-quality proposals (Sec.~\ref{sec:rpn attack}).

	\item \textbf{Step 3}. To enhance the attacking strength further, UPC then jointly attacks RPN as well as the classification and the bounding box regression head by lowering the detection scores and distorting the bounding box (Sec.~\ref{sec:c&r attack}). 
\end{itemize}
We perform these steps in an iterative manner until the termination criterion is satisfied, \ie, fooling rate is larger than the threshold or the iteration reaches the maximum. 

\paragraph{Attacking in Physical Space.}~
By imposing the semantic constraint (Sec.~\ref{section3.4}), the generated camouflage patterns by UPC look natural for human observers and thus can be regarded as texture patterns on human accessories. Concretely, we pre-define several regions of human accessories (\eg, garment, mask) to paint on the generated camouflage patterns (Fig.~\ref{fig:pattern_plan}) for attacking, and the corresponding physical scenes are captured under different viewing conditions (\eg, illumination, viewpoints) for testing (Fig.~\ref{fig:pipeline}(b)).

\subsection{Physical Simulation}
\label{section3.4}

\paragraph{Material Constraint.} 
To keep generated adversarial patterns less noticeable, the perturbations are camouflaged as texture patterns on human accessories (\eg, garment, mask). External environments are simulated via controlling factors such as lighting, viewpoint, location and angle~\cite{chen2018shapeshifter,evtimov2017robust}. To effectively handle non-rigid or non-planar objects,
we also introduce addition transformation functions to model their internal deformations (Eq.~\ref{eqn:generate_perturbation}).

\paragraph{Semantic Constraint.}
Inspired by the imperceptibility constraint in digital attacks, we use the projection function (Eq.~\ref{eqn:project_perturbation}) to enforce the generated adversarial patterns to be visually similar to natural images during optimization. Empirical results show that optimizing with this constraint yields high-quality semantic patterns, which can be naturally treated as camouflages on human clothing (Fig.~\ref{fig:patterns}). 

\paragraph{Training Data.} 
To obtain universal patterns, images with different human attributes (body sizes, postures, \etc) are sampled as the training set $\mathcal{\mathcal{X}}$.

In summary, the perturbed images are generated by:
\vspace{-.5em}
\begin{equation} \label{eqn:project_perturbation}
\delta^{t} = Proj_{\infty}(\delta^{t-1}+\Delta \delta,I,\epsilon),
\end{equation}
\vspace{-1.5em}
\begin{equation} \label{eqn:generate_perturbation}
\mathcal{\hat{\mathcal{X}}}=\left\{ \hat{x_i}|\hat{x_i}=T_r(x_i+T_c(\delta^{t})),x_i \sim \mathcal{X} \right\}.
\end{equation}
Eq.~\ref{eqn:project_perturbation} is the semantic constraint, where $\delta^{t}$ and $\Delta \delta$ denote the adversarial pattern and its updated vector at iteration $t$, respectively.
$Proj_{\infty}$ projects generated pattern onto the surface of $L_\infty$ norm-balls with radius $\epsilon$ and centered at $I$. Here we choose $I$ as natural images to ensure the generated camouflage patterns are semantically meaningful.
Eq.~\ref{eqn:generate_perturbation} is the physical simulation we applied during the attack, where $T_r$ is applied to all training images and used for the environmental simulation (\eg, illumination).
 $T_c$ is acted on generated patterns, which is used for modeling the material constraint (\eg, deformations induced by stretching). $\hat{x}$ is the generated perturbed image (marked as blue in Fig.~\ref{fig:pipeline}(a)).

\subsection{Region Proposal Network (RPN) Attack}
\label{sec:rpn attack}
For an input image with height $H$ and width $W$, the RPN extracts $M=O(HW)$ proposals across all anchors.
We denote the output proposals of each image $\hat{x}$ as $\mathcal{P}=\left\{p_i|p_i=(s_i,\vec{d_i});i=1,2,3...M \right\}$,
where $s_i$ is the confidence score of $i$-th bounding box and
$\vec{d_i}$ represents the coordinates of $i$-th bounding box.
We define the objective function for attacking the RPN as following:
\vspace{-.25em}
\begin{equation} \label{eqn:rpn_loss}
L_{rpn}= \mathop {\mathbb{E}} \limits_{p_i\sim \mathcal{P} } ( \mathcal{L}(s_i,y^t) + s_i \| \vec{d_i}-\Delta \vec{d_i}\|_p ),
\vspace{-0.5em}
\end{equation}
where $y^t$ is the target score, and we set $y^{1}$ for background and $y^{0}$ for foreground;
$\mathcal{L}$ is the Euclidean distance loss;
$\Delta \vec{d_i}$ is a pre-difined vector, which used for attacking proposals by shifting the center coordinate and corrupting the shape of original proposals;
$p$ is the norm constant and we set $p=1$ in the experiment.

By minimizing $L_{rpn}$, our goal is to generate adversarial patterns for RPN which results in a substantial reduction of foreground proposals and severely distorted candidate boxes (marked as red in Fig.~\ref{fig:pipeline}(a)).

\subsection{Classifier and Regressor Attack}
\label{sec:c&r attack}
After applying non-maximum suppression (NMS) on the outputs of RPN, top-$k$ proposals are ordered by their confidence scores and selected as a subset $\mathcal{\hat{P}}$. These top-scored proposals $\mathcal{\hat{P}}$ are then fed to the classification and the regression head for generating final outputs.
We note that if only a subset of proposed bounding boxes are perturbed,
the detection result of the attacked image may still be correct if a new set of candidate boxes is picked in the next iteration, which results in great challenges for attackers.
To overcome this issue, we instead extract proposals densely as in \cite{xie2017adversarial}.
Specifically, we attack an object by either decreasing the confidence of the groundtruth label or increasing the confidence of the target label.
We further enhance the attacking strength by distorting the aspect ratio of proposals and shifting the center coordinate simultaneously~\cite{li2018robust}. In summary, we attack the classification and the regression head by:
\vspace{-.25em}
\begin{equation}\label{eqn:cls_loss}
\begin{split}
L_{cls}=\mathop {\mathbb{E}} \limits_{p\sim \mathcal{\hat{P}}} C(p)_{y} + \mathop {\mathbb{E}} \limits_{p\sim \mathcal{P^{*}}}\mathcal{L}(C(p),y') ,
\end{split}
\end{equation}
\vspace{-1.35em}
\begin{equation}\label{eqn:reg_loss}
\begin{split}
L_{reg}=\sum_{p\sim \mathcal{P^{*}}} \|R(p)_y-\Delta \vec{d}\|_l ,
\end{split}
\vspace{-1.75em}
\end{equation}
where $\mathcal{L}$ is the cross-entropy loss, $C$ and $R$ are the prediction output of the classifier and the regressor. $\mathcal{P^{*}}$ is the proposals which can arybe detected as true label $y$, and $y'$ is the target label for attacking. 
$\Delta \vec{d}$ denotes the distortion offset. We select $\ell_{2}$ norm, \ie, $l=2$ in Eq.~\ref{eqn:reg_loss}.
Eq.~\ref{eqn:cls_loss} and  Eq.~\ref{eqn:reg_loss} are designed for fooling the classifier and the regressor, respectively, and are referred to as C\&R attack (marked as green in Fig.~\ref{fig:pipeline}(a)).
For \textbf{untargeted attack}, we set $y=y'$ for maximizing (instead of minimizing) Eq.~\ref{eqn:cls_loss}.

\begin{algorithm}[t!]
	\small
	\caption{Algorithm of UPC}
	\begin{algorithmic}[1]
		\Require
		Training images $\mathcal{X}$;
		Target label $y'$;
		Balance parameters $\lambda_1$, $\lambda_2$;
		Iteration parameters $iter_{s}$ and $iter_{max}$;
		Fooling rate threshold $r_s$;
		\Ensure
		Universal adversarial pattern $\delta$;
		Fooling rate $r$;
		\State $\delta^{0} \gets random$, $\Delta\delta \gets 0$, $r \gets 0$, $t \gets 0$
		\While {$t<iter_{max}$ and $r<r_s$}
		\State $t \gets t+1$, $\delta^{t} \gets Proj_{\infty}(\delta^{t-1}+\Delta \delta,I,\epsilon)$
		\ForAll{$x_i \sim \mathcal{X}$}
		\State Choose the transformation of $T_r$ and $T_c$ randomly
		\State $\hat{x_i}=clip~(T_r(x_i+T_c(\delta^{t})),0,1)$
		\EndFor
        \State Caculate the fooling rate $r$ of perturbed images $\hat{\mathcal{X}}$
		\If {$t<iter_{s}$ and $r<r_s$}
		\State $\mathop {argmin} \limits_{\Delta\delta}\mathop {\mathbb{E}} \limits_{\hat{x_i} \sim \mathcal{\hat{X}} } L_{rpn} +L_{tv}$
		\Else
		\State $\mathop {argmin} \limits_{\Delta\delta}\mathop {\mathbb{E}} \limits_{\hat{x_i} \sim \mathcal{\hat{X}} } (L_{rpn}+ \lambda_1L_{cls} +  \lambda_2L_{reg})+L_{tv}$
		\EndIf
		\EndWhile
		\label{code:recentEnd}
	\end{algorithmic}
	\label{Algorithm}
\end{algorithm}

\begin{figure*}[t]
\begin{center}
\includegraphics[width=\linewidth]{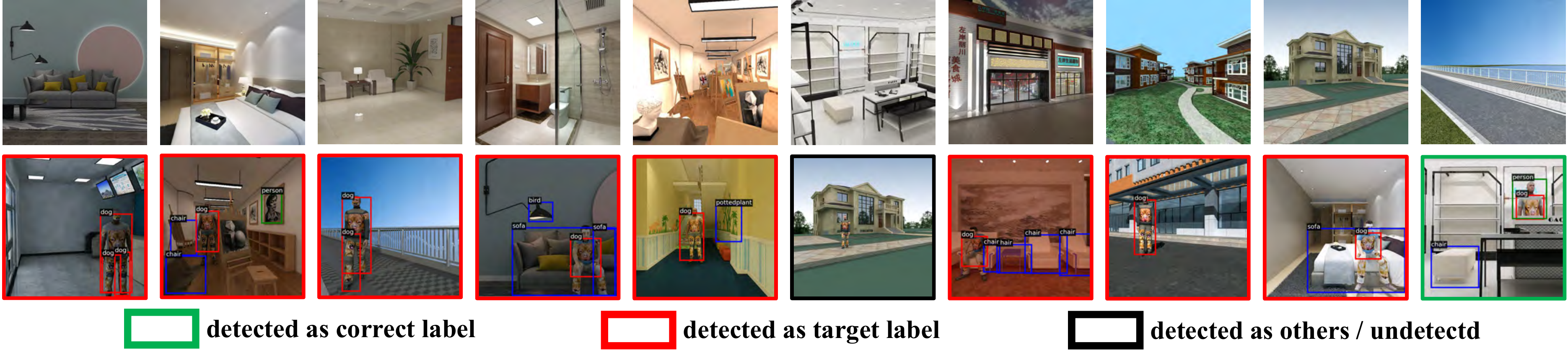}
\vspace{-2.1em}
\end{center}
   \caption{
    \textbf{Examples of virtual scene experiments.}
    Virtual scenes (\ie, \textbf{\emph{AttackScenes}}) are shown in the first row, including indoors and outdoors environments.
	The second rows shows results captured under various physical conditions with different pattern schemes.
   }
\vspace{-1em}
\label{fig:scene}
\end{figure*}

\subsection{Two-Stage Attacking Procedure}
In summary, UPC generates the physical universal adversarial perturbations by considering all the factors above:
\vspace{-.5em}
\begin{equation}\label{eqn:total_loss}
\mathop {argmin} \limits_{\Delta \delta}
\mathop {\mathbb{E}} \limits_{\hat{x} \sim \mathcal{\hat{\mathcal{X}}}}
(L_{rpn}+ \lambda_1L_{cls} + \lambda_2L_{reg}) + L_{tv}(\delta^{t}) ,
\end{equation}
where $\delta$ and $\hat{\mathcal{X}}$ denote the universal pattern and the set of perturbed images, respectively. $L_{tv}$ stands for the \textbf{total variation loss}~\cite{mahendran2015understanding} with \textbf{$\ell_{2}$} norm constraint applied. We note that $L_{tv}$ is important for reducing noise and producing more natural patterns.

The overall procedure of UPC is illustrated in Algorithm~\ref{Algorithm}, where we alternately update the universal perturbation pattern $\delta$ and the perturbed images $\hat{x} \sim \mathcal{\hat{\mathcal{X}}}$ until the fooling rate becomes larger than a certain threshold or the attack iteration reaches the maximum.
$\delta$ is updated using a two-stage strategy.
During the first stage, we exclusively attack the RPN to reduce the number of valid proposals, \ie, set $\lambda_1=0$ and $\lambda_2=0$ in Eq.~\ref{eqn:total_loss}.
After significantly reducing the number of high-quality proposals, 
our attack then additionally fools the classification and bounding box regression head in the second stage. By minimizing Eq.~\ref{eqn:total_loss}, the generated perturbation $\delta$ substantially lowers the quality of proposals and thereby achieves a high fooling rate.

%% file: 4_evaluation.tex
\begin{figure}[t]
\begin{center}
\includegraphics[width=0.85\linewidth]{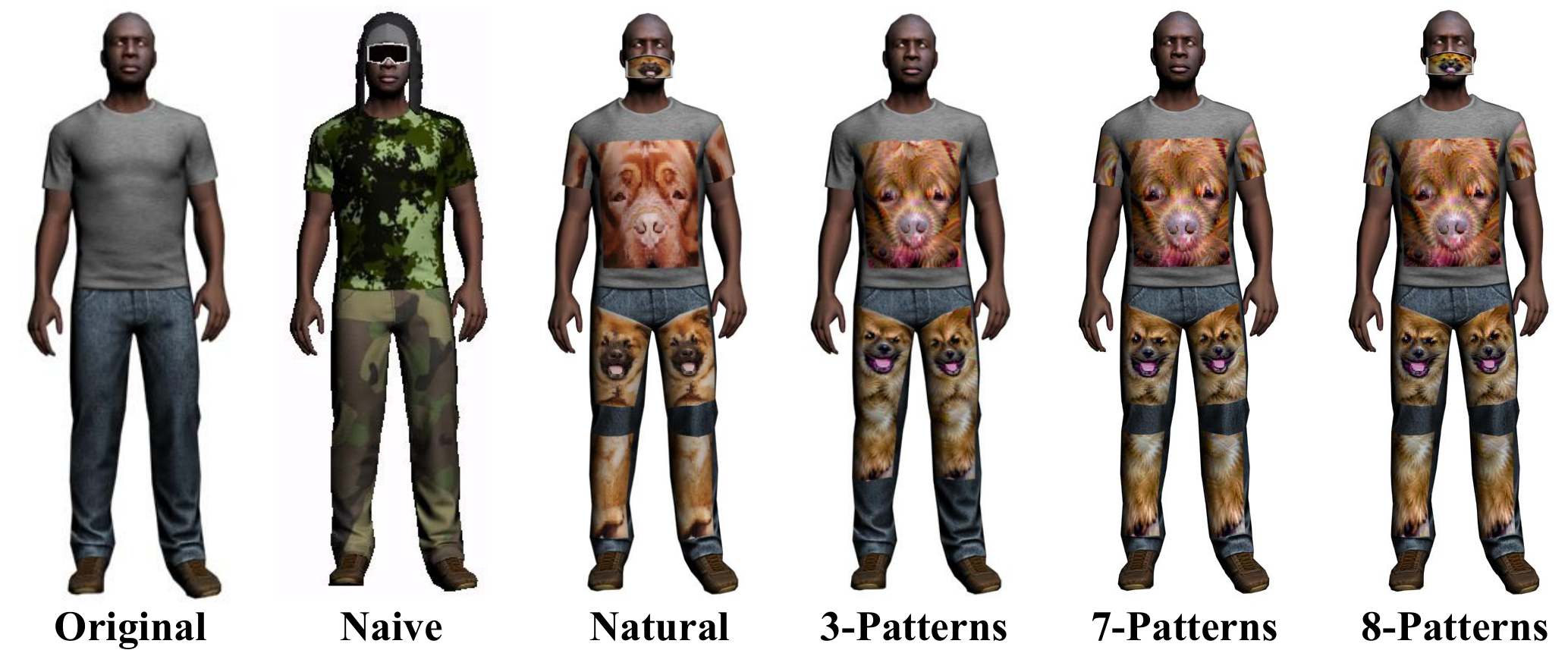}
\end{center}
\vspace{-1.2em}
   \caption{
    \textbf{Examples of pattern schemes in the virtual scenes experiment.} 		\textbf{Original:} humans without camouflage patterns;
		\textbf{Naive:} humans with simple camouflages (\ie, army camouflage cloths, pilot cap and snow goggles);  
		\textbf{Natural:} humans with natural images as camouflage patterns. 
		\textbf{3/7/8-Patterns:} according to the heatmaps of detection models, we pre-define 3/7/8 regions on human accessories to paint on the generated camouflage patterns.
   }
\vspace{-1em}
\label{fig:pattern_plan}
\end{figure}

\section{AttackScenes Dataset}
Due to the lack of a standardized benchmark dataset, earlier works measure the performance under irreproducible physical environments, which makes it difficult to make fair comparisons between different attacks.
To this end, we build the first standardized dataset, named \emph{\textbf{AttackScenes}}, for fair and reproducible  evaluation.

\paragraph{Environments.}
\emph{\textbf{AttackScenes}} includes 20 virtual scenes under various physical conditions (Fig.~\ref{fig:scene}).
Specifically, there are 10 indoors scenes (\eg, bathroom, living room) 
and 10 outdoors scenes(\eg, bridge, market) in total.

\paragraph{Camera Setting.}
For each virtual scene, 18 cameras are placed for capturing images from different viewpoints.
To ensure the diversity of images, these cameras are located at different angles, heights and distances (Fig.~\ref{fig:pipeline}(b)).

\paragraph{Illumination Control.}
To the best of our knowledge, earlier studies usually conduct tests in bright environments. However, this simulated condition is quite limited since there exist many dark scenes in the real world.
Accordingly, we extend the testing environment to better simulate different daily times like evening and dawn.
Area lights and directional light sources are used to simulate indoors and outdoors illuminations, respectively.
The illumination varies from dark to bright at 3 levels by controlling the strength of light sources (\ie, L1$\sim$L3).

\section{Experiments}
In this section, we empirically show the effectiveness of the proposed UPC by providing thorough evaluations in both virtual and physical environments.

\subsection{Implementation Details}
We mainly evaluate the effectiveness of our method on ``person'' category due to its importance in video surveillance and person tracking~\cite{li2018scale}.
We collect 200 human images with various attributes (\eg, hair color, body size) as our training set to generate universal adversarial patterns. Following \cite{xie2017adversarial}, we evaluate the performance of faster r-cnn
using 2 network architectures (\ie, VGG-16~\cite{simonyan2014very} and ResNet-101\cite{he2016deep})
which are either trained on the PascalVOC-2007 \texttt{trainval}, or on the combined set of PascalVOC-2007 \texttt{trainval} and PascalVOC-2012 \texttt{trainval}.
We denote these models as FR-VGG16-07, FR-RES101-07, FR-VGG16-0712 and FR-RES101-0712.

\paragraph{Parameters setting.}
We set fooling rate threshold $r_s=0.95$, $iter_{s}=100$ and the maximum iteration $iter_{max}=2000$ in Algorithm \ref{Algorithm}. 
More parameters and transformation details are recorded in sec.~1 of supplementary material.

\paragraph{Evaluation Metric.} 
For faster r-cnn, we set the threshold of NMS as 0.3 and the confidence threshold as 0.5 (instead of the default value 0.8).
Even though IoU is used for standard evaluation of object detection, we do not use this metric here since our focus is whether the detector hits or misses the true label of the attacked instance. To this end, we extend the metrics in \cite{chen2018shapeshifter,evtimov2017robust} to be applicable in our experiments, 
precision $p_{0.5}$, to measure the probability of whether the detector can hit the true category:
\vspace{-0.5em}
\begin{equation}
\vspace{-0.5em}
	p_{0.5} = \frac{1}{{|\mathcal{X}|}}\sum_{v\sim\mathbb{V},b\sim\mathbb{B},s\sim\mathbb{S}} \left\{ \mathop {C(x)} \limits_{x\in\mathcal{X}} =y, \mathop {C(\hat{x})} \limits_{\hat{x}\in\hat{\mathcal{X}}} =y \right\},
\end{equation}

where $x$ is the original instance and $\hat{x}$ denotes the instance with camouflage patterns. $\mathbb{V},\mathbb{L},\mathbb{S}$ denote the sets of camera viewpoints, brightness and scenes, respectively;
$C$ is the prediction of detector and $y$ is the groundtruth label (\ie, person, car).

\begin{table}[]
	\scriptsize
	\centering
	\caption{Average precision $p_{0.5}$ in virtual scene experiments after attacking faster r-cnn. Note that $p_{0.5}$ is averaged over all viewpoints of each pattern scheme under 3 brightness conditions.}
	\resizebox{\linewidth}{!}{
		\begin{tabular}{c|cccc|cccc}
			\hline
			Network                       &\multicolumn{4}{c|}{FR-VGG16-0712}  &\multicolumn{4}{c}{FR-RES101-0712}\\
			\thickhline
			\multirow{2}{*}{Schemes}          &\multicolumn{4}{c|}{Standing}      &\multicolumn{4}{c}{Standing}  \\ \cline{2-9}
			& L1 &L2 &L3  & Avg (Drop)                    & L1 &L2 &L3 & Avg (Drop)                     \\ \cline{1-9}
			Original                       &  0.97    & 0.97      & 1.0       &  0.98 (-)               &   0.99    & 0.99      & 1.0       &  0.99 (-)  \\
			Naive                       &  0.97    & 0.97      & 0.99      &  0.97 (0.01)             &  0.99    & 0.99      & 0.99      &  0.99 (0.0)\\
			Natural                       &  0.95    & 0.96      & 0.98      &  0.96 (0.02)                &  0.97    & 0.97      & 0.98      &  0.97 (0.02)\\
			3-Patterns      &  0.64    & 0.36      & 0.18      &  0.39 (\textbf{0.59})     &  0.73    & 0.69      & 0.70      &  0.69 (\textbf{0.30})  \\ 
			7-Patterns                &  0.55    & 0.33      & 0.22      &  0.37 (\textbf{0.61})   &  0.51    & 0.48      & 0.64      &  0.54 (\textbf{0.45})  \\ 
			8-Patterns                   &  0.15    & 0.03      & 0.02      &  0.07 (\textbf{0.91})  &  0.10    & 0.09      & 0.13      &  0.11 (\textbf{0.88})  \\ \hline
			\multirow{2}{*}{Schemes}          &\multicolumn{4}{c|}{Walking}      &\multicolumn{4}{c}{Walking}  \\ \cline{2-9}
			& L1 &L2 &L3  & Avg (Drop)                    & L1 &L2 &L3 & Avg (Drop)                     \\ \cline{1-9}
			Original                &  0.93    & 0.94      & 0.99      &  0.95 (-)             &  0.98    & 0.99       & 1.0      &  0.99 (-)  \\	
			Naive                 &  0.92    & 0.94      & 0.96      &  0.94 (0.01)           &  0.98    & 0.97      & 0.98      &  0.98 (0.01)  \\
			Natural                &  0.91    & 0.93      & 0.95      &  0.93 (0.02)             &  0.98    & 0.99      & 0.98      &  0.98 (0.01) \\
			3-Patterns           &  0.37    & 0.26      & 0.16      &  0.26 (\textbf{0.69})      &  0.44    & 0.50      & 0.50      &  0.48 (\textbf{0.51})  \\ 
			7-Patterns            &  0.28    & 0.25      & 0.16      &  0.23 (\textbf{0.72})    &  0.31    & 0.33      & 0.34      &  0.33 (\textbf{0.66})  \\
			8-Patterns                &  0.06    & 0.05      & 0.01      &  0.04 (\textbf{0.91})    &  0.05    & 0.06      & 0.06      &  0.06 (\textbf{0.93})  \\\hline
			\multirow{2}{*}{Schemes}          &\multicolumn{4}{c|}{Sitting}      &\multicolumn{4}{c}{Sitting}  \\ \cline{2-9}
			& L1 &L2 &L3  & Avg (Drop)                    & L1 &L2 &L3 & Avg (Drop)                     \\ \cline{1-9}
			Original                       &  0.97    & 0.99      & 0.99      &  0.98  (-)                &  1.0     & 0.99      & 0.99      &  0.99  (-)   \\
			Naive                    &  0.93    & 0.94      & 0.95      &  0.94 (0.04)              &  0.93    & 0.92      & 0.93      &  0.93 (0.06)  \\
			Natural             &  0.94    & 0.94      & 0.98      &  0.95 (0.03)            &  0.97    & 0.98      & 0.98      &  0.98 (0.01) \\
			3-Patterns                &  0.83    & 0.64      & 0.63      &  0.70 (\textbf{0.28})     &  0.75    & 0.77      & 0.79      &  0.77
			(\textbf{0.22})  \\
			7-Patterns                  &  0.83    & 0.77      & 0.63      &  0.74 (\textbf{0.24})     &  0.77    & 0.78      & 0.78      &  0.78 (\textbf{0.21}) \\
	        8-Patterns                 &  0.60    & 0.47      & 0.32      &  0.46 (\textbf{0.52})    &  0.49 & 0.57  & 0.62 & 0.56 (\textbf{0.43})  \\
        \hline
		\end{tabular}
	}
	\label{table:virtual_experiment}
	\vspace{-1em}
\end{table}

\subsection{Virtual Scene Experiment}
\label{sec:virtual}
\paragraph{Human Model and Pattern Schemes.}~
We select human models in \emph{\textbf{AttackScenes}} with different poses (\ie, standing, walking and sitting) as the attacking target.
6 different schemes ({Fig.~\ref{fig:pattern_plan}}) are used under the material constraint (Sec.~\ref{section3.4}) for experimental comparison.

\paragraph{Comparison Between Pattern Schemes.}
In the virtual scene experiment, $1080(20\times3\times18)$ images are rendered for each pattern scheme.
Without loss of generality, we choose ``dog'' and ``bird'' as target labels to fool detectors in our experiment.
We use 6 different pattern schemes illustrated in Fig.~\ref{fig:pattern_plan}
for validating the efficacy of the proposed UPC.

As shown in Table~\ref{table:virtual_experiment}, we find that the attack strength is generally weaker in darker environments. This can be attributed to the fact that the adversarial patterns are badly captured when the level of brightness is low, which induces low-quality attacks.
Additionally, we observe that for different human poses the average precision almost stays at the same level
via attacking Naive/Natural pattern scheme which indicates that simply using naive camouflage or natural images as adversarial patterns is invalid for physical attacks. 
By contrast, our method yields a distinct drop rate of $p_{0.5}$ for all 3 pattern schemes (\ie, 3/7/8-Pattern schemes), among which 8-Pattern scheme observes the highest performance drop (\ie, Standing: $p_{0.5}$ drops from $0.98$ to $0.07$ 
using FR-VGG16). 
It is no surprise to observe such a phenomenon since using more generated patterns for physical attack results 
leads to a higher fooling rate.
The detection result further shows our attack is invariant to different viewing conditions (\eg, viewpoints, brightness). Additionally, we also find that among these 3 poses ``Sitting'' is the most difficult to attack since some patterns (\eg, pants or cloth patterns) are partially occluded (see sampled images from Fig.~\ref{fig:example} and Fig.~\ref{fig:scene}). 

\begin{table}[t!]
	\centering
	\scriptsize
	\setlength{\tabcolsep}{4mm}
	\caption{ Performance comparison with prior arts of physical attacks under different settings. We record $p_{0.5}$ and drop rate averaged over all viewpoints of 8-pattern scheme.}
	\begin{tabular}{lccc}
		\toprule
		Network & \multicolumn{3}{c}{FR-VGG16-0712}  \\
		\toprule
		Setup                &  Standing 	&      Walking   &  Sitting \\
		\midrule
		$UPC_{rc}$ \textbf{(ours)}              &  0.07 (\textbf{\underline{0.91}})   & 0.04 (\textbf{\underline{0.91}})     & 0.46 (\textbf{\underline{0.52}})  \\
		$UPC_{r}$ \textbf{(ours)}             &  0.66 (0.32)       & 0.33 (0.62)           &  0.76 (0.22)  \\
		$CLS_{rc}$ \textbf{(ours)}               & 0.18 (0.80)      & 0.06 (0.89)            &   0.54 (0.44)      \\
		$Shape~\cite{chen2018shapeshifter}$               &  0.70 (0.28)     &  0.39 (0.56)      &  0.78 (0.20)     \\
		$ERP^2~\cite{evtimov2017robust}$               & 0.85 (0.13)      & 0.48 (0.47)            &   0.87 (0.11)      \\
		$AdvPat~\cite{thys2019fooling}$               & 0.77 (0.21)      & 0.31 (0.64)            &   0.78 (0.20)      \\
		\bottomrule	
		\toprule
		Network & \multicolumn{3}{c}{FR-RES101-0712}  \\
		\toprule
		Setup                &  Standing 	&      Walking   &  Sitting \\
		\midrule
		$UPC_{rc}$ \textbf{(ours)}              &  0.11 (\textbf{\underline{0.88}})    &  0.06 (\textbf{\underline{0.93}})    & 0.56 (\textbf{\underline{0.43}}) \\
		$UPC_{r}$ \textbf{(ours)}             &0.73 (0.26)        & 0.42 (0.57)              &  0.86 (0.13) \\
		$CLS_{rc}$ \textbf{(ours)}               &0.30 (0.69)     &0.16 (0.83)            &   0.65 (0.34)     \\
		$Shape~\cite{chen2018shapeshifter}$               &0.83 (0.16)      &0.47 (0.52)      &   0.88 (0.11)     \\
		$ERP^2~\cite{evtimov2017robust}$               & 0.79 (0.20)    & 0.44 (0.55)           & 0.91 (0.08)      \\
		$AdvPat~\cite{thys2019fooling}$               & 0.91 (0.08)      & 0.71 (0.28)            &   0.93 (0.06)      \\
		\bottomrule			
	\end{tabular}
	\vspace{-2em}
	\label{table:compare_table}
\end{table}

\paragraph{Compare with Existing Attacks.}
We compare UPC with existing physical attacks under the following settings (Table~\ref{table:compare_table}): (1) both internal deformations $T_c$ and external physical environments $T_r$ are simulated in Eq.~\ref{eqn:generate_perturbation}, denoted as $UPC_{rc}$; (2) only external physical environments are modeled, \ie, $T_r$ is used in Eq.~\ref{eqn:generate_perturbation}, denoted as $UPC_{r}$.
(3) only attack the classification head, \ie, $L_{cls}$ is used to generate patterns, denoted as $CLS_{rc}$; 
(4) ShapeShifter~\cite{chen2018shapeshifter}, \ie, only use $T_r$ in Eq.~\ref{eqn:generate_perturbation} and attack against the classifier, denoted as $Shape$. (5) we follow~\cite{song2018physical} by extending $RP^2$~\cite{evtimov2017robust} for attacking faster r-cnn, denoted as $ERP^2$, and (6) Adversarial Patches~\cite{thys2019fooling}, which utilize various transformations to fool all proposals across images, denote as $AdvPat$. 
These six scenarios were tested under same training setup (detailed in sec.1 of supplementary material).

The performance of 8-patterns scheme is recorded in Table~\ref{table:compare_table}, and the implications are two-fold. First, we can see the drop rates of $p_{0.5}$ in $UPC_{rc}$ and $CLS_{rc}$ are significantly higher than those of $UPC_{r},SS$ and $ERP^2$. 
These quantitative results indicate that the proposed transformation function $T_c$ can effectively mimic the deformations (\eg, stretching) of complex objects. Second, $UPC_{rc}$ and $UPC_{r}$ outperform $CLS_{rc}$ and $Shape$, which suggest that the joint attack paradigm (\ie, RPN and C\&R attack) generally shows stronger attacking strength than only attacking the classification head~\cite{chen2018shapeshifter}. In conclusion, all these experimental results demonstrate the efficacy of the proposed transformation term $T_c$ as well as the joint attack paradigm for fooling object detectors in the wild. Moreover, our proposed UPC outperforms existing methods~\cite{chen2018shapeshifter, evtimov2017robust, thys2019fooling}, and thereby establish state-of-the-art for physical adversarial attack on proposal-based object detectors.

The visualization of discriminative regions are showed in supplementary material~\cite{selvaraju2017grad}. 
We can observe that the UPC has superior attacking capability while other methods can not depress the activated features of un-occluded parts effectively, which may lead higher detection accuracy.

\begin{figure}[t]
	\begin{center}
		\includegraphics[width=0.97\linewidth]{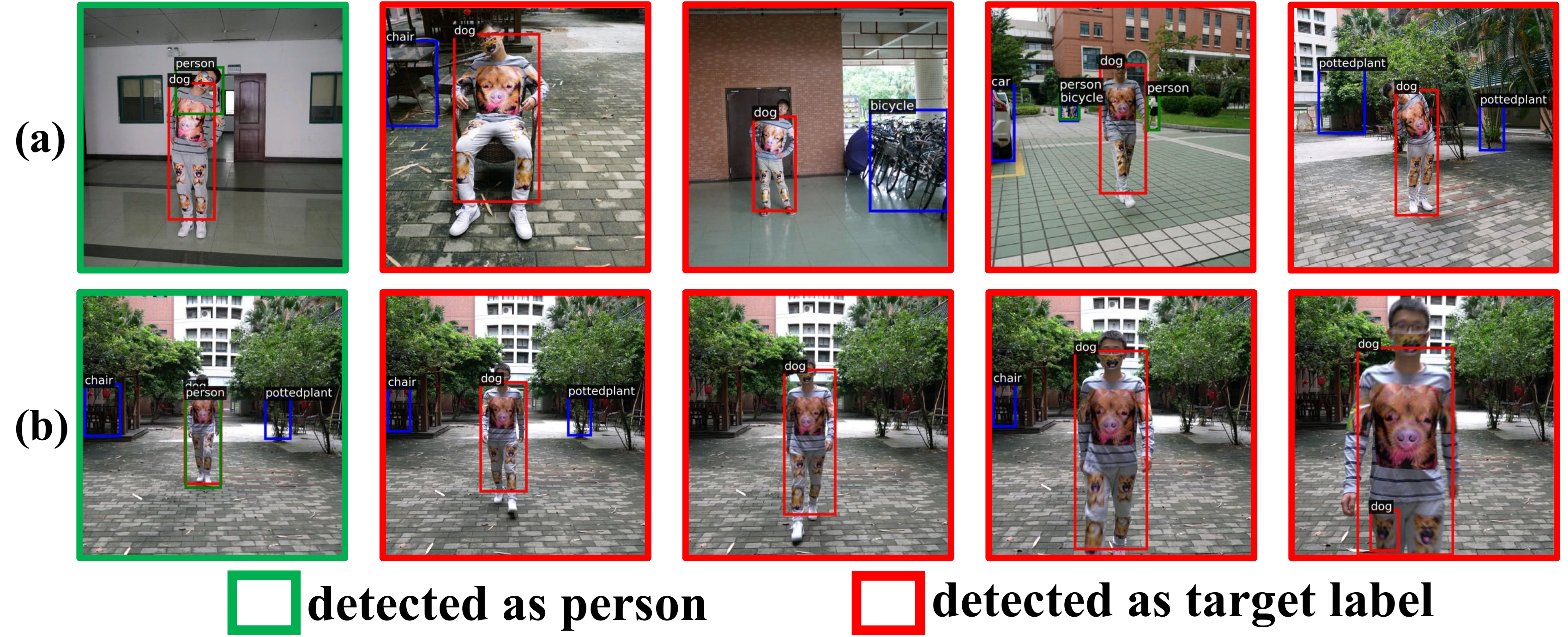}
	\end{center}
	\vspace{-1em}
	\caption{
	\textbf{Experimental results in (a) stationary testing and (b) motion testing.}
		The camouflage is generated by FR-VGG16. 
	}
	\label{fig:physical}
	\vspace{-1.5em}
\end{figure}

\subsection{Physical Environment Experiment}
\label{sec:physical environment}
Following the setup of virtual scene experiments, we stick the same camouflage pattern on different volunteers with diverse body sizes and garment styles.
During the physical experiment, we use Sony$\alpha7r$ camera to take photos and record videos.
Our physical experiments include two parts: \textbf{stationary testing} and \textbf{motion testing}.

\paragraph{Stationary Testing.}
In the physical world, we choose 5 scenes including indoors and outdoors scenes under different lighting conditions.
Similar to virtual scene experiments, we take 18 photos of the attacked person for each pattern scheme.
To evaluate the robustness of our method under different deformations, the person is required to switch from 6 different poses (\ie, standing, sitting, leg lifting, waving hands, fork waist, shaking head) during photographing (Fig.~\ref{fig:physical}(a)).
We record the average precision $p_{0.5}$ and drop rates of FR-VGG16-0712 and FR-RES101-0712 under three brightness conditions in Table~\ref{table:stationary} (detailed in sec.2 of supplementary material).
Similar to our findings in~Sec.~\ref{sec:virtual}, UPC expresses its superior attacking capability 
in the real physical world compared to natural image patterns which results in nearly zero drop rate in every posture.

As can be seen from Table~\ref{table:virtual_experiment} and Table~\ref{table:stationary}, the behaviors of detectors exhibit similar trends
under different physical conditions such as lighting conditions in both virtual scenes and physical environments. Another noteworthy comment is that the generated patterns from virtual scene experiments demonstrate high transferability to the real physical world (Table~\ref{table:stationary}).
These facts indicate that our \emph{\textbf{AttackScenes}} is a suitable dataset to study physical attacks.

\begin{table}[]
	\scriptsize
	\centering
	\caption{Average precision $p_{0.5}$ and drop rate under 3 brightness conditions in stationary testing.}
	\resizebox{\linewidth}{!}{
		\begin{tabular}{c|cc|cc}
			\hline
			Network                       &\multicolumn{2}{c|}{FR-VGG16-0712}  &\multicolumn{2}{c}{FR-RES101-0712}\\
			\thickhline
			Schemes          &\multicolumn{1}{c}{Standing}      &\multicolumn{1}{c|}{Shaking Head}                  &\multicolumn{1}{c}{Standing}   &\multicolumn{1}{c}{Shaking Head}  \\ \cline{1-5}
			Original                       & 1.0 (-)  &  1.0 (-)  & 1.0  &  1.0 (-)     \\
			Natural                       & 0.98 (0.02)  & 0.98 (0.02) & 0.98 (0.02)  &  1.0 (0.0)      \\
			3-Patterns                &  0.67 (\textbf{0.33}) & 0.74 (\textbf{0.26}) & 0.72 (\textbf{0.28}) &  0.76 (\textbf{0.24})   \\
			7-Patterns                   & 0.59 (\textbf{0.41}) & 0.59 (\textbf{0.41}) & 0.59 (\textbf{0.41}) &  0.57 (\textbf{0.43}) \\
			8-Patterns                   &  0.17 (\textbf{0.83}) &  0.20 (\textbf{0.80}) & 0.19 (\textbf{0.81}) &  0.20 (\textbf{0.80}) \\
			\hline
			Schemes          &\multicolumn{1}{c}{Fork Waist}      &\multicolumn{1}{c|}{Leg Lifting}                  &\multicolumn{1}{c}{Fork Waist}   &\multicolumn{1}{c}{Leg Lifting}  \\ \cline{1-5}
			Original                       & 1.0 (-)  & 1.0 (-)  & 1.0 (-)  &  1.0 (-)     \\
			Natural                       &  1.0 (0.0)  & 1.0 (0.0) &  1.0 (0.0)  &   1.0 (0.0)   \\
			3-Patterns                & 0.72 (\textbf{0.28}) & 0.74 (\textbf{0.26}) & 0.76 (\textbf{0.24}) & 0.71 (\textbf{0.29})   \\
			7-Patterns                   & 0.56 (\textbf{0.44}) & 0.54 (\textbf{0.46}) & 0.57 (\textbf{0.43}) &  0.57 (\textbf{0.43}) \\
			8-Patterns                   & 0.20 (\textbf{0.80}) & 0.26 (\textbf{0.74}) & 0.24 (\textbf{0.76}) &  0.30 (\textbf{0.70}) \\ \hline
			Schemes          &\multicolumn{1}{c}{Rasing Hands}      &\multicolumn{1}{c|}{Sitting}                  &\multicolumn{1}{c}{Rasing Hands}   &\multicolumn{1}{c}{Sitting}  \\ \cline{1-5}
			Original                       & 1.0 (-)  & 1.0 (-) & 1.0 (-)  &  1.0 (-)     \\
			Natural                       & 0.98 (0.02)  &  1.0 (0.0) &  1.0 (0.0)  &   1.0 (0.0)      \\
			3-Patterns                & 0.83 (\textbf{0.17}) & 0.76 (\textbf{0.24}) & 0.85 (\textbf{0.15}) &  0.74 (\textbf{0.26})   \\
			7-Patterns                   & 0.65 (\textbf{0.35}) & 0.54 (\textbf{0.46}) & 0.69 (\textbf{0.31}) &  0.59 (\textbf{0.41}) \\ 
			8-Patterns                   & 0.35 (\textbf{0.65}) & 0.22 (\textbf{0.78}) & 0.35 (\textbf{0.65}) &  0.26 (\textbf{0.74}) \\ \hline
		\end{tabular}
	}
	\label{table:stationary}
	\vspace{-1em}
\end{table}

\paragraph{Motion Testing.}

To further demonstrate the efficacy of UPC, we also test our algorithm on human motions. 
The video clips were obtained under different physical conditions (\eg, different lighting conditions, scenes) while the volunteers are walking towards the camera. Meanwhile, they are randomly changing postures from the 6 classes as mentioned above. A total of 3693 frames where 583,  377, 219, 713, 804 and 997 frames are collected under 5 different physical scenes so as to make this dataset diverse and representative. And the detection precisions are $26\%$ (150/583), $21\%$ (80/377), $17\%$ (37/219), $34\%$ (240/713), $15\%$ (118/804) and $ 24\%$ (240/997), respectively. Experiments in all physical scenes have observed low detection rates, which further confirms the effectiveness of UPC. The detection results of some sampled frames are shown in Fig.~\ref{fig:physical}(b), where people are detected as ``dog''. 
We find this attack is much more effective under brighter conditions. This phenomenon coincides with previous observations in virtual scene studies (Sec.~\ref{sec:virtual}), and also further justify the potential value of \emph{\textbf{AttackScenes}}. Moreover, we find that blurred camouflage patterns during motion make UPC less effective, which lead to higher detection accuracy. 

\begin{figure}[t]
\begin{center}
\includegraphics[width=\linewidth]{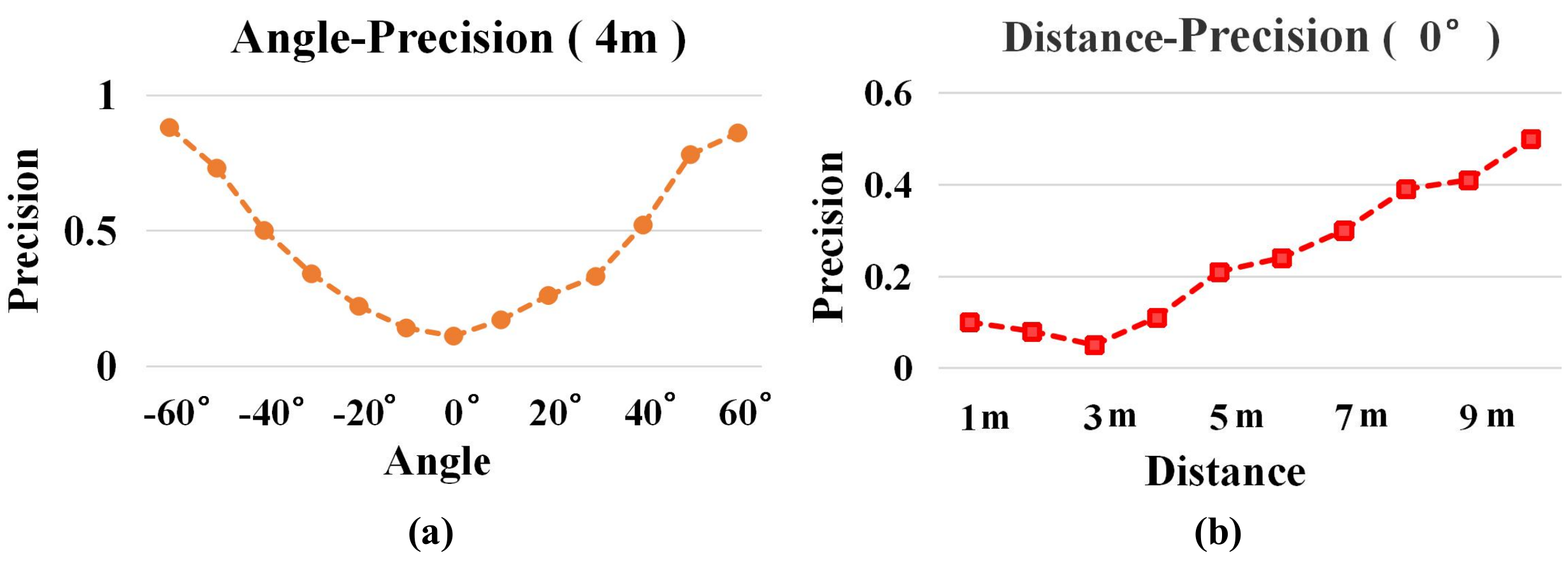}
\end{center}
\vspace{-1.5em}
   \caption{
    \textbf{The precision $p_{0.5}$ of detectors under different angle/distance conditions}.
    We note that high viewing angle or far distance can make attacks less effective.
   }
\label{fig:data}
\vspace{-1.2em}
\end{figure}

We also plot the relationship between the detection precision \emph{vs.} angle/distance under 8-Pattern schemes as in Fig.~\ref{fig:data}. It can be concluded that when the absolute value of the angle/distance between the person and the camera becomes larger, camouflage patterns are captured with lower quality and thus hampering the attacks.

\subsection{Transferability Experiment}
We generate camouflage patterns from one architecture to attack other models.
In our experiment, FR-VGG16-0712 and FR-RES101-0712 are used to compute camouflage patterns.
We introduce ResNet-50, ResNet-152 and MobileNet~\cite{howard2017mobilenets} based faster r-cnn which are trained on MS-COCO2014~\cite{lin2014microsoft} dataset as transfer-testing models.
Other architecture models including R-FCN (ResNet-101)~\cite{Dai2016R}, SSD (VGG-16)~\cite{Liu2016SSD}, Yolov2~\cite{redmon2017yolo9000}, Yolov3~\cite{redmon2018yolov3} and RetinaNet~\cite{Lin_2017_ICCV} are considered in our transferability experiments.
Eight models are publicly available, and we denote them as
FR-RES50-14, FR-RES152-14, FR-MN-14, RFCN-RES101-07, SSD-VGG16-0712, Yolov2-14, Yolov3-14 and Retina-14.
The confidence threshold of all models is set as 0.5 for evaluation.

The following experiments are conducted: (1) \textbf{Cross-Training Transfer.}
The transferability between source and attacked models have the same architecture but are trained on different datasets (\eg, using the pattern generated from FR-VGG16-0712 to attack FR-VGG16-07);
(2) \textbf{Cross-Network Transfer}.
The transferability through different network structures (\eg, using the pattern computed from FR-VGG16-0712 to attack Yolov3-14).

For transfer experiments, virtual walking humans with 8-Patterns scheme (Fig.~\ref{fig:pattern_plan}) are used to evaluate the transferability under transfer attacks.
The transfer performances are illustrated in Table~\ref{table:table3}.
The original pattern scheme is used to calculate the baseline precision of each model (denoted as ``Original'' in Table~\ref{table:table3}).
We observe the precisions of all detectors have dropped,
which means the generated patterns exhibits well transferability and generality across different models and datasets. It is noteworthy to mention our proposed UPC also successfully breaks 4 state-of-the-art defenses~\cite{liao2018defense, xie2017mitigating, guo2017countering, prakash2018deflecting} (see Supplementary).

\begin{figure}[t]
\begin{center}
\includegraphics[width=0.95\linewidth]{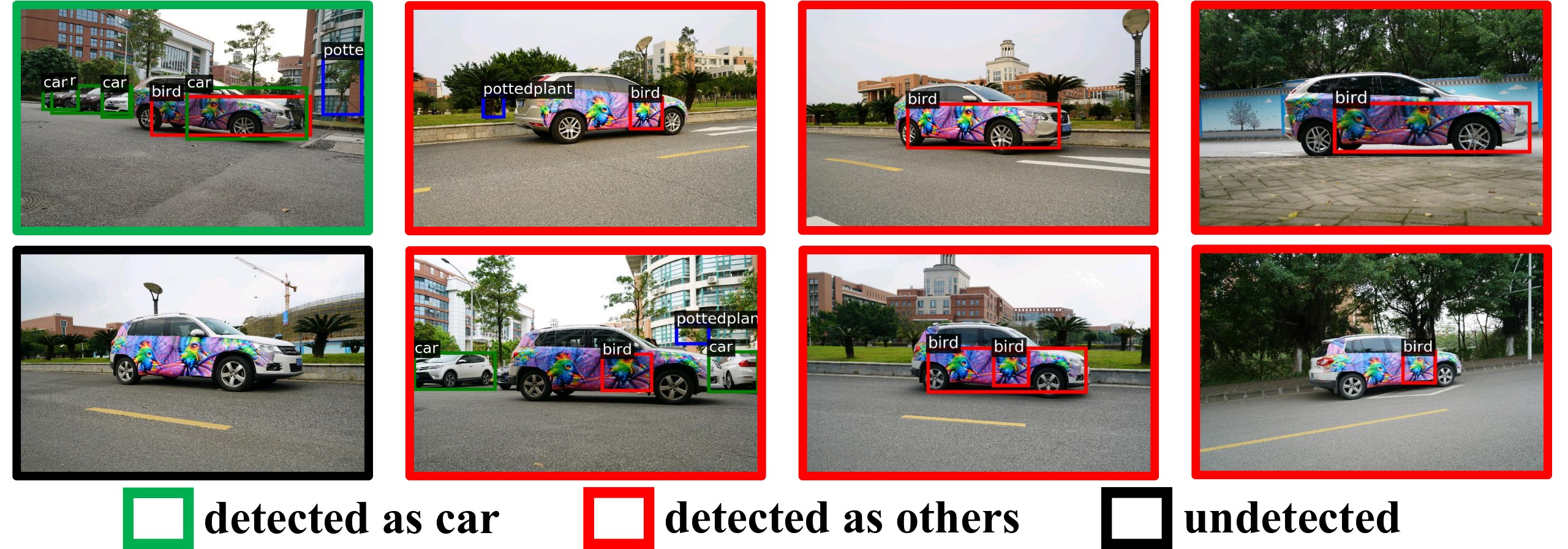}
\end{center}
\vspace{-1.2em}
   \caption{
    \textbf{The results of attacking Volvo XC60 (top row) and Volkswagen Tiguan (bottom row)}.
    The generated camouflage patterns fool detectors to misrecognize the car as bird.
   }
\label{fig:vehicle}
\vspace{-0.5em}
\end{figure}

\begin{table}[t!]
	\centering
	\scriptsize
	\caption{\label{tab:test}Average precision $p_{0.5}$ in transferability testing. First seven rows show the results of cross-training transfer testing, and rest five rows display the cross-network transfer's results (\textbf{bold} in ``Network" column).}
	\begin{tabular}{lccc}
		\toprule
		\multirow{2}{*}{Network}  & \multirow{2}{*}{Original} &  FR-VGG16-0712                 & FR-RES101-0712 \\  \cline{3-4}
		&                           &  Average (Drop)                & Average (Drop)    \\
		\midrule
		FR-VGG16-0712             & 0.95                      & \textbf{0.04 (\underline{0.91})}          & 0.10 (0.85)          \\
		FR-RES101-0712            & 0.99                      & 0.78 (0.21)                   & \textbf{0.06 (\underline{0.93})}              \\
		FR-VGG16-07               & 0.95                      & 0.08 (\underline{0.87})                   & 0.11 (0.84)                          \\
		FR-RES101-07              & 0.99                      & 0.51 (0.48)                   & 0.10 (\underline{0.89})                           \\
		FR-RES50-14               & 1.0                       & 0.85 (0.15)                   & 0.78 (\underline{0.22})                          \\
		FR-RES152-14              & 1.0                       & 0.62 (0.38)                   & 0.43 (\underline{0.57})                           \\
		FR-MN-14                   & 0.99                      & 0.51 (0.48)                   & 0.25 (\underline{0.74})                          \\
		\textbf{RFCN-RES101-07~\cite{Dai2016R}}    & 0.98                      & 0.64 (0.34)                   & 0.41 (\underline{0.57})                         \\
		\textbf{SSD-VGG16-0712~\cite{Liu2016SSD}}    & 0.75                      & 0.13 (\underline{0.62})                   & 0.16 (0.59)                    \\
		\textbf{Yolov2-14~\cite{redmon2017yolo9000}}        & 1.0                       & 0.59 (0.41)                   & 0.38 (\underline{0.62})                      \\
		\textbf{Yolov3-14~\cite{redmon2018yolov3}}        & 1.0                       & 0.69 (\underline{0.31})                   & 0.71 (0.29)                      \\
		\textbf{Retina-14~\cite{Lin_2017_ICCV}}        & 1.0                       & 0.72 (0.31)                   & 0.49 (\underline{0.51})                      \\
		\bottomrule
	\end{tabular}
	\label{table:table3}
	\vspace{-2.2em}
\end{table}

\subsection{Generalization to Other Categories}
To demonstrate the generalization of UPC, we construct camouflage patterns by untargeted attacks to fool the ``car'' category (\ie, rigid but non-planar object).
We use Volvo XC60 (champagne) and Volkswagen Tiguan (white) as the attacking target in the real world.
The pattern will be regarded as car paintings by human observers. In order to not affect driving, we restrict the camouflage coverage regions to exclude windows, lightings, and tires.
We collect 120 photos which includes different distances (8$\sim$12m) and angles (-45$\degree$$\sim$45$\degree$) in 5 different environments (Fig.~\ref{fig:vehicle}).
The video is recorded simultaneously at same angles.
The performance of pure non-camouflage car is $p_{0.5}=1$, while after attacking only 
 $24\%$ (29/120) images and $26\%$ (120/453) frames are detected as ``car'' correctly, which verifies the efficacy of UPC.

%% file: 5_discussion.tex
\begin{figure}[t]
\begin{center}
\includegraphics[width=\linewidth]{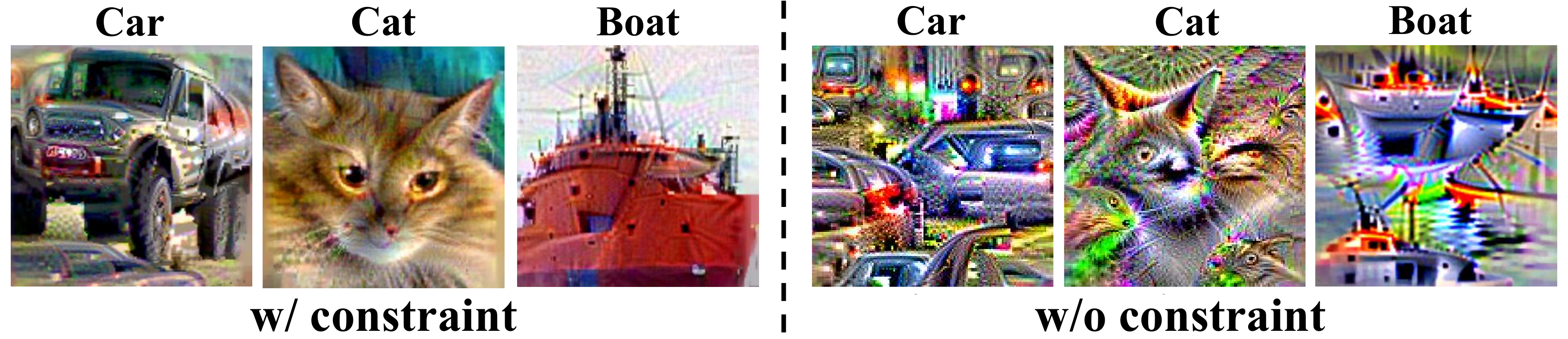}
\end{center}
\vspace{-1em}
   \caption{
    \textbf{Generated camouflage patterns are semantically meaningful}. Even for unconstrained patterns, human observer can relate the generated camouflage patterns to the targeted label.
   }
\vspace{-1em}
\label{fig:patterns}
\end{figure}
\section{Discussion}
\paragraph{Abstract Semantic Patterns.}
A side finding is that the generated patterns without semantic constraint (Eq.~\ref{eqn:project_perturbation}) can be less semantic meaningful but exhibits abstract meanings (Fig.~\ref{fig:patterns}). This observation suggest that human and machine classification of adversarial images are robustly related as suggested in \cite{zhou2019humans}.

\paragraph{Defense Method Evaluation.}
With the development of defense methods in digital domain~\cite{liu2018adv,xie2019feature}, we hope the collected dataset, \emph{\textbf{AttackScenes}}, can benefit future research of defense methods against physical attacks.

%% file: 6_conclusion.tex
\section{Conclusion}
In this paper, we study the problem of physical attacks on object detectors. Specifically, we propose UPC to generate universal camouflage patterns which hide a category of objects from being detected or to misdetect objects as the target label by state-of-the-art object detectors. 
In addition, we present the first standardized benchmark dataset, \emph{\textbf{AttackScenes}}, to simulate the real 3D world in controllable and reproducible environments. This dataset can be used for accessing the performance of physical-world attacks at a fair standard. Our study shows that the learned universal camouflage patterns not only mislead object detectors in the virtual environment, \ie, \emph{\textbf{AttackScenes}}, but also attack detectors successfully in the real world.

\section*{Acknowledgements}
This work was supported by National Key Research and Development Plan in China (2018YFC0830500), National Natural Science Foundation of China under Grant (61972433), Fundamental Research Funds for the Central Universities (19lgjc11, 19lgyjs54).

%% file: supp.tex
\clearpage
\onecolumn
\appendix

\section{Parameter Setup And Transformation Details}
In this section, we provide more details about the parameter setup. We use ADAM optimizer of Tensorflow platform in the attack framework with the default setup: $learning~rate=0.001$, $beta1=0.9$, , $beta2=0.999$ and $epsilon=10^{-8}$. We set $\lambda_1=10.0$ and $\lambda_2=0.1$ in Eq.~6 as initialization. In Algorithm 1, we set  the fooling rate threshold $r_s=0.95$, $iter_{s}=100$ and the maximum iteration $iter_{max}=2000$. 

Total four methods include $UPC$\textbf{(ours)}, $ShapeShifter$, $ERP^2$ and $AdvPat$. FR-RES101-0712/FR-VGG16-0712 are used to generate the patterns for these four methods under same setting (\eg, transformation parameters, training set and training epoch, etc).

\begin{table}[h]
	\small
	\centering
	\setlength{\tabcolsep}{4mm}
	\caption{Distribution of transformations and parameters for $UPC$. }
	\begin{tabular}{ccc|ccc}
		\thickhline
		\multicolumn{6}{c}{$UPC$\textbf{(ours)}}  \\
		\thickhline
		\multicolumn{3}{c|}{$T_r$}                &  \multicolumn{3}{c}{$T_c$} \\
		Transform & Parameters & Remark            &   Transform & Parameters & Remark  \\
		\thickhline
		Affine             & $\mu=0,\sigma=0.1$  & Perspective Transforms   & Affine  & $\mu=0,\sigma=0.03$  & Deformed Simulation \\
		Rotation             &  $-15\degree\sim15\degree$  &  Camera Simulation   & Cropping   & $0.7\sim1.0$   & Occlude Simulation \\
		Contrast           &  $0.5\sim1.5$ & Camera Parameters  & Translation     & $-0.04\sim0.04$ & Pattern Location  \\
		Scale             &  $0.25\sim1.25$  & Distance  & Scale     & $0.95\sim1.05$ & Pattern Size  \\
		Brightness             &  $-0.25\sim0.25$ & Illumination &       &   &  \\
		\thickhline	
	\end{tabular}
	\label{table:UPC_transformation}
	\vspace{-1 em}
\end{table}

\begin{table}[h] 
	\small
	\centering
	\setlength{\tabcolsep}{5mm}
	\caption{Distribution of transformations and parameters for $ShapeShifter$.}
	\begin{tabular}{ccc}
		\thickhline
		\multicolumn{3}{c}{$ShapeShifter$}  \\
		\thickhline
		Transform & Parameters & Remark           \\
		\thickhline
		Translation             &  $-0.2\sim0.2$ & Perspective Transforms    \\
		Rotation             &  $-15\degree\sim15\degree$  & Camera Simulation \\
		Contrast           &  $0.5\sim1.5$ & Camera Parameters        \\
		Scale             &  $0.25\sim1.25$  & Distance   \\
		Brightness             &  $-0.25\sim0.25$ & Illumination \\
		\thickhline	
	\end{tabular}
	\label{table:Shape_transformation}
	\vspace{-1 em}
\end{table}

\begin{table}[h]
	\small
	\centering
	\setlength{\tabcolsep}{5mm}
	\caption{Distribution of transformations and parameters for $ERP^2$.}
	\begin{tabular}{ccc}
		\thickhline
		\multicolumn{3}{c}{$ERP^2$}  \\
		\thickhline
		Transform & Parameters & Remark           \\
		\thickhline
		Affine             &  $\mu=0,\sigma=0.1$   & Perspective Transforms    \\
		Cropping             &  $0.9\sim1.2$  &  Photograph Simulation \\
		Contrast           &  $0.5\sim1.5$ & Camera Parameters        \\
		\thickhline	
	\end{tabular}
	\label{table:ERP}
	\vspace{-1 em}
\end{table}

\begin{table}[h]
	\small
	\centering
	\setlength{\tabcolsep}{5mm}
	\caption{Distribution of transformations and parameters for $AdvPat$.}
	\begin{tabular}{ccc}
		\thickhline
		\multicolumn{3}{c}{$AdvPat$}  \\
		\thickhline
		Transform & Parameters & Remark           \\
		\thickhline
		Random Noise             &  $-0.15\sim0.15$ & Noise    \\
		Rotation             &  $-15\degree\sim15\degree$  &  Camera Simulation \\
		Contrast           &  $0.5\sim1.5$ & Camera Parameters        \\
		Scale             &  $0.8\sim1.2$  & Resize   \\
		Brightness             &  $-0.25\sim0.25$ & Illumination \\
		\thickhline	
	\end{tabular}
	\label{table:AdvPat}
\end{table}

\clearpage

\section{Experiments in Physical World}
\subsection{Result of Stationary Testing}
To evaluate the robustness of our method under different deformations, the person is required to switch from 6 different poses (\ie, standing, sitting, leg lifting, waving hands, fork waist, shaking head) during photographing.
We record the average precision $p_{0.5}$ and drop rates of FR-VGG16-0712 and FR-RES101-0712 under three brightness conditions in Table~\ref{table:stationary_supp}.

\begin{table}[h]
	\tiny
	\centering
	\caption{Average precision $p_{0.5}$ in stationary testing after attacking faster r-cnn. We test on a total of 6 different poses (\ie, standing, sitting, leg lifting, waving hands, fork waist, shaking head).}
	\resizebox{0.8\linewidth}{!}{
		\linespread{1.25}
		\begin{tabular}{c|cccccccc}
			\hline
			Network                       &\multicolumn{8}{c}{FR-VGG16-0712} \\
			\thickhline
			\multirow{2}{*}{Schemes}          &\multicolumn{4}{c}{Standing}      &\multicolumn{4}{c}{Sitting}       \\ \cline{2-9}
			& L1 &L2 &L3  & Avg (Drop)                    & L1 &L2 &L3 & Avg (Drop)            \\ \cline{1-9}
			Original                       & 1.0  & 1.0  & 1.0  &  1.0 (-)               &  1.0  & 1.0   & 1.0  & 1.0 (-)             \\
			Natural                       & 1.0  & 0.94 & 1.0  &  0.98 (0.02)             &  1.0  & 1.0   & 1.0  & 1.0 (0.0)            \\
			3-Patterns                & 0.72 & 0.61 & 0.67 &  0.67 (\textbf{0.33})     &  0.83 & 0.78  & 0.67 & 0.76 (\textbf{0.24})  \\
			7-Patterns                   & 0.67 & 0.56 & 0.56 &  0.59 (\textbf{0.41})  &  0.61 & 0.50  & 0.50 & 0.54 (\textbf{0.46})     \\
			8-Patterns                   & 0.22 & 0.11 & 0.17 &  0.17 (\textbf{0.83})   &  0.28 & 0.17  & 0.22 & 0.22 (\textbf{0.78})               \\ \hline
			\multirow{2}{*}{Schemes}    &\multicolumn{4}{c}{Fork Waist}      &\multicolumn{4}{c}{Leg Lifting}      \\ \cline{2-9}
			& L1 &L2 &L3  & Avg (Drop)                    & L1 &L2 &L3 & Avg (Drop)          \\ \cline{1-9}
			Original                       & 1.0  & 1.0  & 1.0  &  1.0 (-)               &  1.0  & 1.0   & 1.0  & 1.0 (-)            \\
			Natural                 & 1.0    & 1.0   & 1.0    &  1.0 (0.0)             & 1.0  & 1.0  & 1.0  &  1.0 (0.0)          \\
			3-Patterns       &  0.78  & 0.72  & 0.67      &  0.72 (\textbf{0.28})     & 0.72 & 0.78 & 0.72 &  0.74 (\textbf{0.26})         \\
			7-Patterns        &  0.61  & 0.50  & 0.56      &  0.56 (\textbf{0.44})       &0.56 & 0.56 & 0.50  &  0.54 (\textbf{0.46})              \\
			8-Patterns             &  0.28  & 0.17  & 0.17   &  0.20 (\textbf{0.80})   &  0.28 & 0.28 & 0.22 &  0.26 (\textbf{0.74})        \\ \hline
			\multirow{2}{*}{Schemes}          &\multicolumn{4}{c}{Rasing Hands}      &\multicolumn{4}{c}{Shaking Head}              \\ \cline{2-9}
			& L1 &L2 &L3  & Avg (Drop)                    & L1 &L2 &L3 & Avg (Drop) \\ \cline{1-9}
			Original                       & 1.0  & 1.0  & 1.0  &  1.0 (-)               & 1.0    & 1.0   & 1.0    &  1.0 (-)         \\
			Natural                       &  0.94 & 1.0  & 1.0   &  0.98 (0.02)               &  1.0  & 1.0   & 1.0  & 1.0 (0.0)           \\
			3-Patterns                 &0.89 & 0.78  & 0.83 &  0.83 (\textbf{0.17})       &0.78    & 0.78      & 0.67      &  0.74 (\textbf{0.26})            \\
			7-Patterns                &  0.72 & 0.61  & 0.61 &  0.65 (\textbf{0.35})        &0.61    & 0.61      & 0.56      &  0.59 (\textbf{0.41})            )\\
			8-Patterns                  & 0.39 & 0.39 & 0.28  &  0.35 (\textbf{0.65})       & 0.22    & 0.28      & 0.11      &  0.20 (\textbf{0.80})            \\ \hline
			
			\hline
			Network                      &\multicolumn{8}{c}{FR-RES101-0712}\\
			\thickhline
			\multirow{2}{*}{Schemes}          &\multicolumn{4}{c}{Standing}      &\multicolumn{4}{c}{Sitting}   \\ \cline{2-9}
			& L1 &L2 &L3  & Avg (Drop)                    & L1 &L2 &L3 & Avg (Drop)    \\ \cline{1-9}
			Original                    & 1.0    & 1.0   & 1.0    &  1.0 (-) & 1.0    & 1.0   & 1.0    &  1.0 (-)\\
			Natural                       & 0.94    & 1.0   & 1.0    &  0.98 (0.02) & 1.0    & 1.0   & 1.0    &  1.0 (0.0)\\
			3-Patterns                    &  0.83 & 0.67 & 0.67 &  0.72 (\textbf{0.28})    & 0.72 & 0.78  & 0.72 & 0.74 (\textbf{0.26}) \\
			7-Patterns                     & 0.61 & 0.56 & 0.61 &  0.59 (\textbf{0.41})   & 0.61 & 0.67  & 0.50 & 0.59 (\textbf{0.41})\\
			8-Patterns                    & 0.22    & 0.22   & 0.11  &  0.19 (\textbf{0.81}) & 0.28    & 0.22  & 0.22    &  0.26 (\textbf{0.74})\\ \hline
			\multirow{2}{*}{Schemes}    &\multicolumn{4}{c}{Fork Waist}      &\multicolumn{4}{c}{Leg Lifting}             \\ \cline{2-9}
			& L1 &L2 &L3  & Avg (Drop)                    & L1 &L2 &L3 & Avg (Drop)    \\ \cline{1-9}
			Original                      & 1.0    & 1.0   & 1.0    &  1.0 (-) & 1.0    & 1.0   & 1.0    &  1.0 (-)\\
			Natural                & 1.0    & 1.0   & 1.0    &  1.0 (0.0) & 1.0    & 1.0   & 1.0    &  1.0 (0.0)\\
			3-Patterns                    &  0.83  & 0.72  & 0.72      &  0.76 (\textbf{0.24})     & 0.67 & 0.78 & 0.67 &  0.71 (\textbf{0.29})\\
			7-Patterns                     &  0.61  & 0.56  & 0.56      &  0.57 (\textbf{0.43})       &0.67 & 0.50 & 0.56  &  0.57 (\textbf{0.43})\\
			8-Patterns                     &  0.28  & 0.22  & 0.22   &  0.24 (\textbf{0.76})   &  0.33 & 0.33 & 0.22 &  0.30 (\textbf{0.70})\\ \hline
			\multirow{2}{*}{Schemes}          &\multicolumn{4}{c}{Rasing Hands}      &\multicolumn{4}{c}{Shaking Head}              \\ \cline{2-9}
			& L1 &L2 &L3  & Avg (Drop)                    & L1 &L2 &L3 & Avg (Drop)     \\ \cline{1-9}
			Original                       & 1.0    & 1.0   & 1.0    &  1.0 (-) & 1.0    & 1.0   & 1.0    &  1.0 (-)\\
			Natural                  & 1.0    & 1.0   & 1.0    &  1.0 (0.0) & 1.0    & 1.0   & 1.0    &  1.0 (0.0)\\
			3-Patterns                        &0.83 & 0.89  & 0.83 &  0.85 (\textbf{0.15})       &0.72    & 0.78      & 0.78      &  0.76 (\textbf{0.24})\\ 
			7-Patterns                        &  0.89 & 0.61  & 0.56 &  0.69 (\textbf{0.31})        &0.56    & 0.61      & 0.56      &  0.57 (\textbf{0.43})\\
			8-Patterns                       & 0.39 & 0.33 & 0.33  &  0.35 (\textbf{0.65})       & 0.22    & 0.22      & 0.17      &  0.20 (\textbf{0.80})\\
			\hline
		\end{tabular}
	}
	\label{table:stationary_supp}
\end{table}
\clearpage

\subsection{Qualitative Samples of Physical Experiments}
\begin{figure*}[h]
	\begin{center}
		\includegraphics[width=0.75\linewidth]{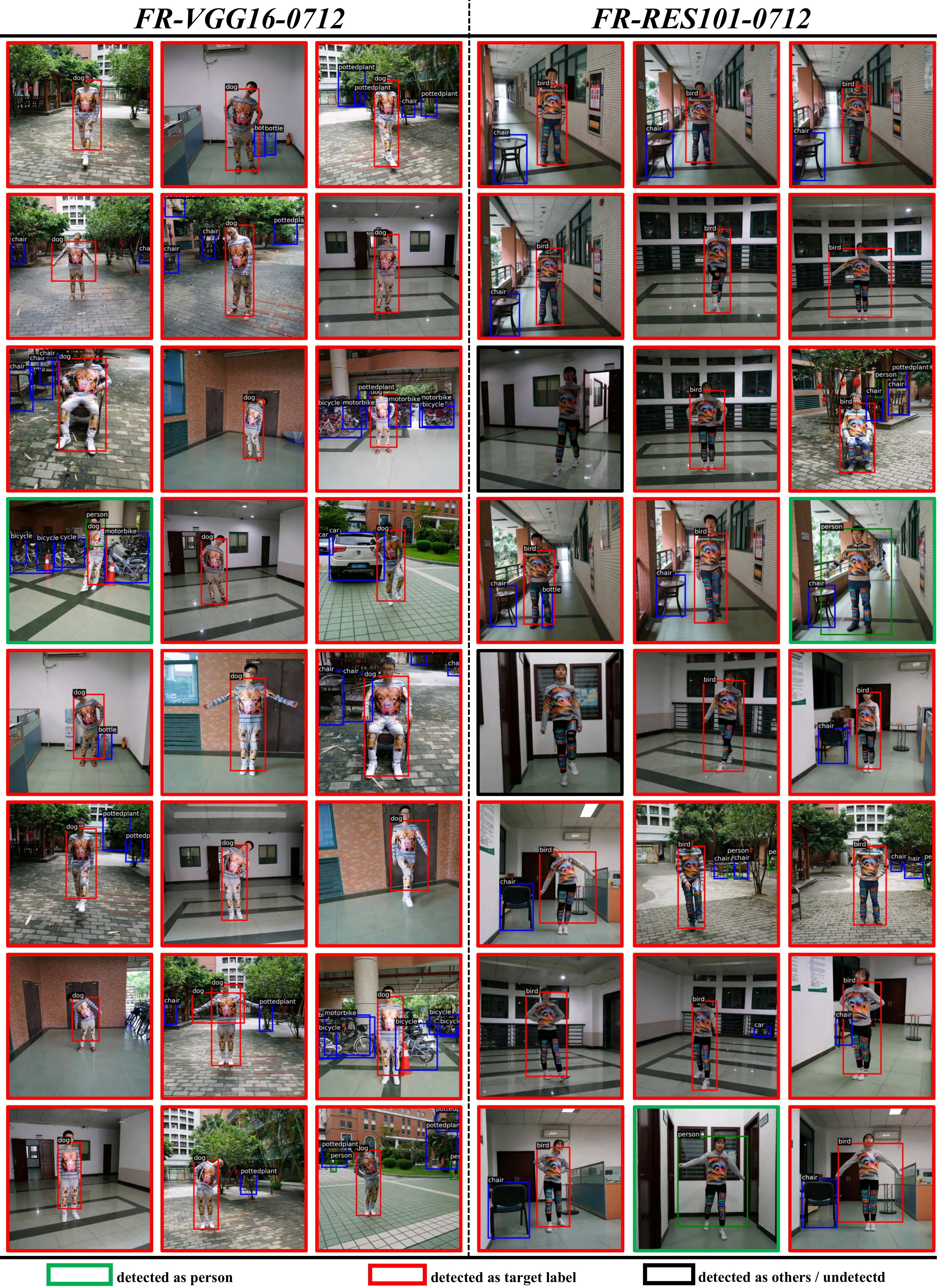}
	\end{center}
	\vspace{-0.5em}
	\caption{
		\textbf{More qualitative results of FR-VGG16-0712 and FR-RES101-0712 on in physical environment.}
		These universal camouflage patterns are generated using FR-VGG16-0712 and FR-RES101-0712, respectively.
		Each row applies different pattern schemes (\ie, 8/7/3-Pattern schemes), and captured in different viewpoints and background environments.
	}
	\label{fig:physical_scenes}
	\vspace{-0.5em}
\end{figure*}

In this section, we provide more qualitative results of FR-VGG16-0712 and FR-RES101-0712 in physical environment  in Figure \ref{fig:physical_scenes}. The detection result further shows our attack is invariant to different viewing conditions (\eg, viewpoints, brightness).

\clearpage

\section{Experiments for Defense Methods}
\begin{figure*}[h]
	\begin{center}
		\includegraphics[width=0.8\linewidth]{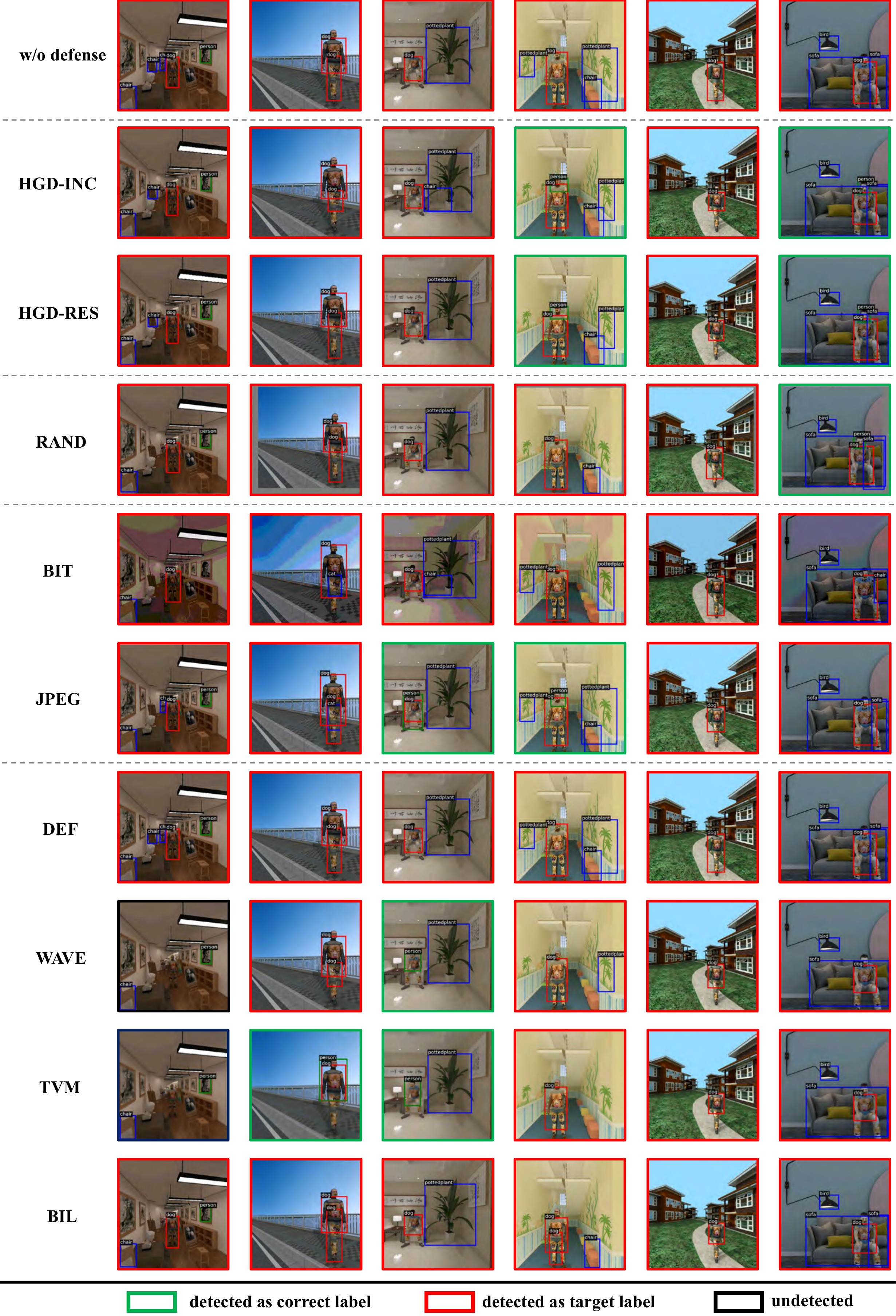}
	\end{center}
	\vspace{-0.5em}
	\caption{
		\textbf{Sampled qualitative results of experiments for defense methods.} We test four state-of-the-art defense methods (\ie, HGD~\cite{liao2018defense}, Randomization~\cite{xie2017mitigating}, Transformations~\cite{guo2017countering} and Deflection~\cite{prakash2018deflecting}). Both quantitative and qualitative result shows these methods can not defend against UPC effectively.
	}
	\label{fig:defense}
	\vspace{-0.5em}
\end{figure*}

In order to test whether our proposed UPC can evade from defense methods, we introduce four state-of-the-art defense methods (\ie, HGD~\cite{liao2018defense}, Randomization~\cite{xie2017mitigating}, Transformations~\cite{guo2017countering} and Deflection~\cite{prakash2018deflecting}) as attacked target. In our experiment, FR-VGG16-0712 and FR-RES101-0712 are used to compute camouflage patterns. We record the experiment results of 8-Pattern schemes in Table~\ref{table:defense}. The original rendered images are used to calculate the baseline precision of each method (denoted as ``w/o defense'' in Table~\ref{table:defense}). The qualitative results of this experiment are displayed in Figure~\ref{fig:defense}.

We observe that the precisions $\hat{p}_{0.5}$ of all defenses stays at a low level,
which means the our proposed UPC successful breaks these state-of-the-art defend methods. An interesting finding is that some methods (\ie, WAVE, TVM) improve the fooling ability of UPC instead of defending against the attacks (see result of ``Sitting'' column).

\paragraph{HGD~\cite{liao2018defense}.}
We utilize trained model (\ie, HGD-inception-v3, HGD-resnet-152) to denoise the rendered images.

\paragraph{Randomization~\cite{xie2017mitigating}.}
In our experiment, we resize the rendered images to $299\times299\times3$, and then transform the images (\ie, random resizing, zero-padding) as the input feed into the detectors, denote as RAND.

\paragraph{Transformation~\cite{guo2017countering}.}
Here we test two defense method (\ie, bit depth reduction (denoted as BIT) and JPEG compression (denoted as JPEG)). For BIT, we set reduce bits number $n=5$ (\ie, for each pixel value, we reduces last 5 bits). For JPEG, we set compression at quality level $q=0.75$. 

\paragraph{Deflection~\cite{prakash2018deflecting}.}
We consider four different settings: 1) we use the pixel deflection mechanism alone to defend UPC, denoted as DEF ; 2) we combine pixel deflection with wavelet denoising, denoted as WAVE ($\sigma=0.04$); 3) pixel deflection and total variance minimization are combined, denoted as TVM; 4) pixel deflection and bilateral filters are composited, denoted as BIL ($size=5,bins=1000$). We set pixel deflection number $n=2000$ during the experiments.

\paragraph{Evaluation Metrics.}
During the evading experiment, we use a the evaluation metric $\hat{p}_{0.5}$ to evaluate the attack performance:
\vspace{-0.5em}
\begin{equation}
\vspace{-0.5em}
\hat{p}_{0.5} = \frac{1}{{|\mathcal{X}|}}\sum_{v\sim\mathbb{V},b\sim\mathbb{B},s\sim\mathbb{S}} \left\{ \mathop {C(x)} \limits_{x\in\mathcal{X}} =y, \mathop {C(\mathcal{D}(\hat{x}))} \limits_{\hat{x}\in\hat{\mathcal{X}}} =y \right\},
\end{equation}
where $x$ is normal person and $\hat{x}$ denotes person with camouflage pattern, $\mathcal{D}$ is the defense method we mentioned before, $\mathbb{V},\mathbb{L},\mathbb{S}$ denote camera viewpoints, brightness and scenes, respectively;
$C$ is the prediction of detector and $y$ is the true label.

\linespread{1.25}
\begin{table}[h]
\tiny
\centering
\caption{Average precision $p_{0.5}$ of defense methods for attacking faster r-cnn. We test on a total of 4 different method (\ie, HGD~\cite{liao2018defense}, Randomization~\cite{xie2017mitigating}, Transformation~\cite{guo2017countering} and Deflection~\cite{prakash2018deflecting}).}

\resizebox{0.95\linewidth}{!}{
	\begin{tabular}{cc|cccccccccccc}
		\hline
		\multicolumn{2}{c}{Network}                       &\multicolumn{12}{|c}{FR-VGG16-0712} \\
		\thickhline
		\multicolumn{2}{c}{\multirow{2}{*}{Defend Method} }         &\multicolumn{4}{|c}{Standing}      &\multicolumn{4}{c}{Walking}   &\multicolumn{4}{c}{Sitting}    \\ \cline{3-14}
		
		\multicolumn{2}{c}{} & \multicolumn{1}{|c}{L1} &L2 &L3  & Avg (Drop)                    & L1 &L2 &L3 & Avg (Drop)       & L1 &L2 &L3 & Avg (Drop)      \\ \cline{2-14}
		
		\multicolumn{2}{c}{w/o defense} & \multicolumn{1}{|c}{0.15}    & 0.03      & 0.02      &  0.07 (\textbf{0.91})                  &  0.06    & 0.05      & 0.01      &  0.04 (\textbf{0.91})       &  0.60    & 0.47      & 0.32      &  0.46 (\textbf{0.52})      \\ \cline{2-14}
		
		\multicolumn{2}{c}{HGD-INC~\cite{liao2018defense}} & \multicolumn{1}{|c}{0.38}    & 0.20      & 0.02      &  0.20 (\textbf{0.78})                  &  0.11    & 0.05      & 0.03      &  0.06 (\textbf{0.89})       &  0.70    & 0.59      & 0.51      &  0.60 (\textbf{0.38})      \\ 
		
		\multicolumn{2}{c}{HGD-RES~\cite{liao2018defense}} & \multicolumn{1}{|c}{0.40}    & 0.22      & 0.02      &  0.21 (\textbf{0.77})                  &  0.11    & 0.05      & 0.02      &  0.06 (\textbf{0.89})       &  0.74    & 0.60      & 0.54      &  0.63 (\textbf{0.35})      \\ \cline{2-14}
		
		\multicolumn{2}{c}{RAND~\cite{xie2017mitigating}} & \multicolumn{1}{|c}{0.28}    & 0.07      & 0.01      &  0.12 (\textbf{0.86})                  &  0.09    & 0.05      & 0.01      &  0.05 (\textbf{0.90})       &  0.63    & 0.51      & 0.40      &  0.51 (\textbf{0.47})      \\ \cline{2-14}
		
		\multicolumn{2}{c}{BIT~\cite{guo2017countering}} & \multicolumn{1}{|c}{0.56}    & 0.28      & 0.0      &  0.28 (\textbf{0.70})                  &  0.26    & 0.04      & 0.03      &  0.11 (\textbf{0.84})       &  0.50    & 0.50      & 0.32      &  0.44 (\textbf{0.54})      \\ 
		
		\multicolumn{2}{c}{JPEG~\cite{guo2017countering}} & \multicolumn{1}{|c}{0.40}    & 0.14      & 0.02      &  0.17 (\textbf{0.81})                  &  0.10    & 0.09     & 0.01      &  0.07 (\textbf{0.88})       &  0.67    & 0.56      & 0.38      &  0.54 (\textbf{0.44})      \\ \cline{2-14}
		
		\multicolumn{2}{c}{DEF~\cite{prakash2018deflecting}} & \multicolumn{1}{|c}{0.22}    & 0.05      & 0.02      &  0.10 (\textbf{0.88})                  &  0.06    & 0.07      & 0.02      &  0.05 (\textbf{0.90})       &  0.60    & 0.47      & 0.32      &  0.46 (\textbf{0.52})      \\ 
					
		\multicolumn{2}{c}{WAVE~\cite{prakash2018deflecting}} & \multicolumn{1}{|c}{0.18}    & 0.13      & 0.0      &  0.11 (\textbf{0.87})                  &  0.14    & 0.07      & 0.02      &  0.08 (\textbf{0.87})       &  0.13    & 0.19      & 0.21      &  0.18 (\textbf{0.80})      \\ 
								
		\multicolumn{2}{c}{TVM~\cite{prakash2018deflecting}} & \multicolumn{1}{|c}{0.60}    & 0.40      & 0.01      &  0.34 (\textbf{0.64})                  &  0.23    & 0.13      & 0.02      &  0.13 (\textbf{0.82})       &  0.57    & 0.57      & 0.45      &  0.53 (\textbf{0.45})      \\ 
											
		\multicolumn{2}{c}{BIL~\cite{prakash2018deflecting}} & \multicolumn{1}{|c}{0.30}    & 0.18      & 0.01      &  0.16 (\textbf{0.82})                  &  0.12    & 0.11      & 0.05      &  0.09 (\textbf{0.86})       &  0.70    & 0.55      & 0.45      &  0.56 (\textbf{0.42})      \\ \cline{2-14}			
		\thickhline

		\multicolumn{2}{c}{Network}                       &\multicolumn{12}{|c}{FR-RES101-0712} \\
		\thickhline
		\multicolumn{2}{c}{\multirow{2}{*}{Defend Method} }         &\multicolumn{4}{|c}{Standing}      &\multicolumn{4}{c}{Walking}   &\multicolumn{4}{c}{Sitting}    \\ \cline{3-14}
		
		\multicolumn{2}{c}{} & \multicolumn{1}{|c}{L1} &L2 &L3  & Avg (Drop)                    & L1 &L2 &L3 & Avg (Drop)       & L1 &L2 &L3 & Avg (Drop)      \\ \cline{2-14}
		
		\multicolumn{2}{c}{w/o defense} & \multicolumn{1}{|c}{0.10}    & 0.09      & 0.13      &  0.11 (\textbf{0.88})                  &  0.05    & 0.06      & 0.06      &  0.06 (\textbf{0.93})       &  0.49    & 0.57      & 0.62      &  0.56 (\textbf{0.43})      \\ \cline{2-14}
		
		\multicolumn{2}{c}{HGD-INC~\cite{liao2018defense}} & \multicolumn{1}{|c}{0.33}    & 0.24      & 0.30      &  0.29 (\textbf{0.70})                  &  0.03    & 0.03      & 0.04      &  0.03 (\textbf{0.96})       &  0.50    & 0.61      & 0.55      &  0.55 (\textbf{0.44})      \\ 
		
		\multicolumn{2}{c}{HGD-RES~\cite{liao2018defense}} & \multicolumn{1}{|c}{0.29}    & 0.23      & 0.24      &  0.25 (\textbf{0.74})                  &  0.04    & 0.04      & 0.03      &  0.04 (\textbf{0.95})       &  0.50    & 0.65      & 0.56      &  0.57 (\textbf{0.42})      \\ \cline{2-14}
		
		\multicolumn{2}{c}{RAND~\cite{xie2017mitigating}} & \multicolumn{1}{|c}{0.27}    & 0.27      & 0.26      &  0.27 (\textbf{0.72})                  &  0.10    & 0.08      & 0.05      &  0.08 (\textbf{0.91})       &  0.50    & 0.61      & 0.64      &  0.58 (\textbf{0.41})      \\ \cline{2-14}
		
		\multicolumn{2}{c}{BIT~\cite{guo2017countering}} & \multicolumn{1}{|c}{0.50}    & 0.12      & 0.08      &  0.23 (\textbf{0.76})                  &  0.19    & 0.04      & 0.03      &  0.9 (\textbf{0.90})       &  0.40    & 0.48      & 0.45      &  0.44 (\textbf{0.55})      \\ 
		
		\multicolumn{2}{c}{JPEG~\cite{guo2017countering}} & \multicolumn{1}{|c}{0.25}    & 0.13      & 0.19      &  0.19 (\textbf{0.80})                  &  0.05    & 0.04     & 0.01      &  0.03 (\textbf{0.96})       &  0.44    & 0.51      & 0.50      &  0.48 (\textbf{0.51})      \\ \cline{2-14}
		
		\multicolumn{2}{c}{DEF~\cite{prakash2018deflecting}} & \multicolumn{1}{|c}{0.13}    & 0.05      & 0.15      &  0.11 (\textbf{0.87})                  &  0.04    & 0.03      & 0.04      &  0.04 (\textbf{0.95})       &  0.45    & 0.57      & 0.58      &  0.54 (\textbf{0.45})      \\ 
		
		\multicolumn{2}{c}{WAVE~\cite{prakash2018deflecting}} & \multicolumn{1}{|c}{0.12}    & 0.16      & 0.20      &  0.16 (\textbf{0.83})                  &  0.11    & 0.01      & 0.04      &  0.05 (\textbf{0.94})       &  0.03   & 0.12      & 0.21      &  0.12 (\textbf{0.87})      \\ 
		
		\multicolumn{2}{c}{TVM~\cite{prakash2018deflecting}} & \multicolumn{1}{|c}{0.50}    & 0.25      & 0.18      &  0.31 (\textbf{0.68})                  &  0.20    & 0.08      & 0.02      &  0.10 (\textbf{0.89})       &  0.17    & 0.38      & 0.33      &  0.29 (\textbf{0.70})      \\ 
		
		\multicolumn{2}{c}{BIL~\cite{prakash2018deflecting}} & \multicolumn{1}{|c}{0.49}    & 0.27      & 0.35      &  0.37 (\textbf{0.62})                  &  0.18    & 0.15      & 0.07      &  0.13 (\textbf{0.86})       &  0.64    & 0.75      & 0.74      &  0.71 (\textbf{0.28})      \\		
		\thickhline								
	\end{tabular}
	}
	\label{table:defense}
\end{table}
\clearpage

\clearpage

\section{Experiments in Virtual Scenes}
Here we provide more qualitative results of FR-VGG16-0712 and FR-RES101-0712 in the synthesized virtual environments.
\subsection{Qualitative Samples of Existing Methods}
\begin{figure*}[h]
	\begin{center}
		\includegraphics[width=0.9\linewidth]{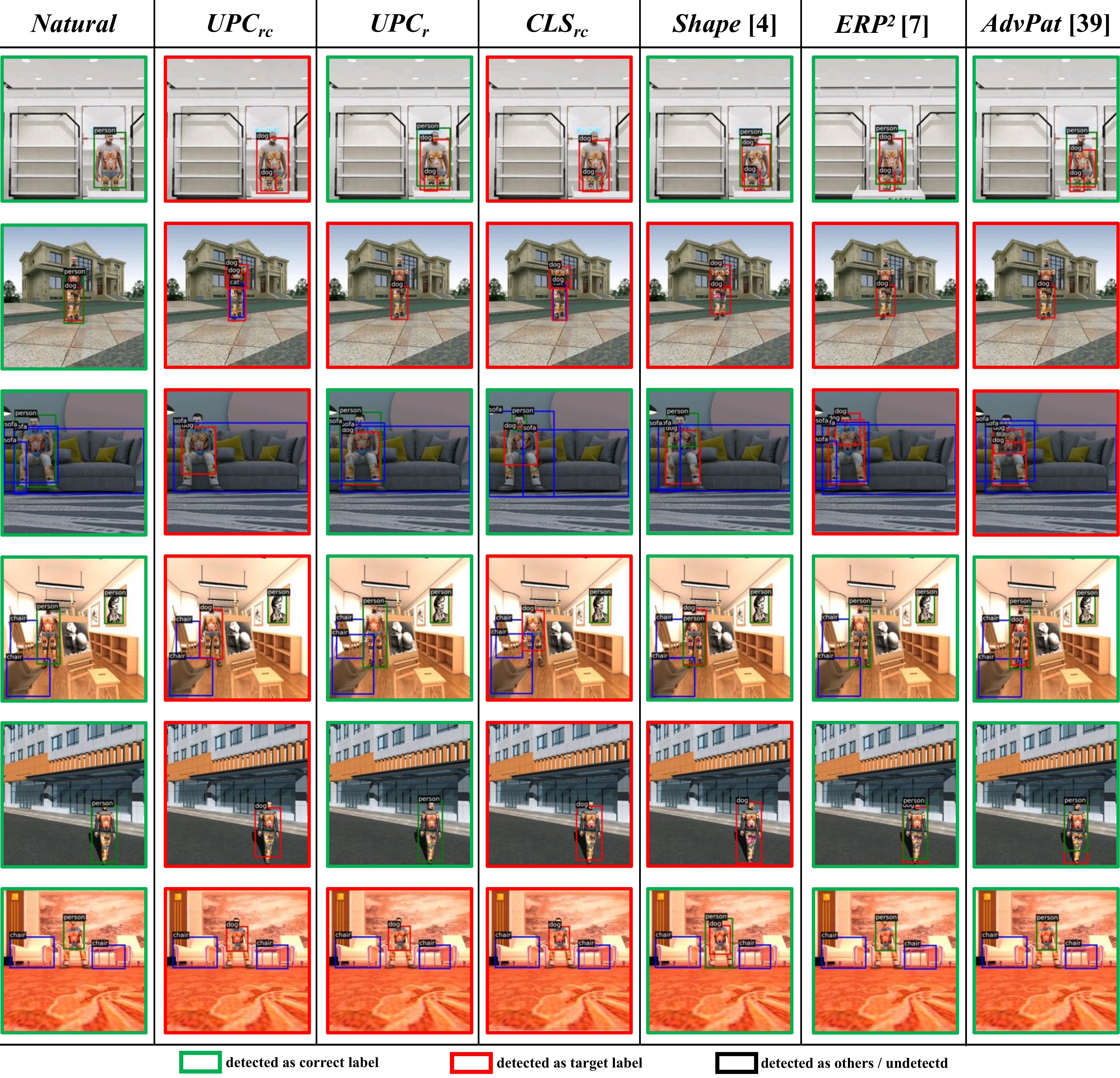}
	\end{center}
	\vspace{-0.50em}
	\caption{
		\textbf{More qualitative results under different attack settings in virtual experiments.} Each column uses same physical conditions (\ie, lighting, viewpoints, environment, \etc). The camouflage patterns generated from $UPC_{rc}$ achieve the most superior performance and visually similar to natural image, which can be regarded as pattern designs on human accessories.
	}
	\label{fig:method_compare}
	\vspace{-0.5em}
\end{figure*}

In this section, we demonstrate several qualitative results of different methods (\ie, $Shape$, $ERP^2$, $AdvPat$ and our proposed $UPC$) in Figure \ref{fig:method_compare}.

\clearpage

\subsection{Detection Result of Various Physical Conditions}
\begin{figure*}[h]
	\begin{center}
		\includegraphics[width=0.8\linewidth]{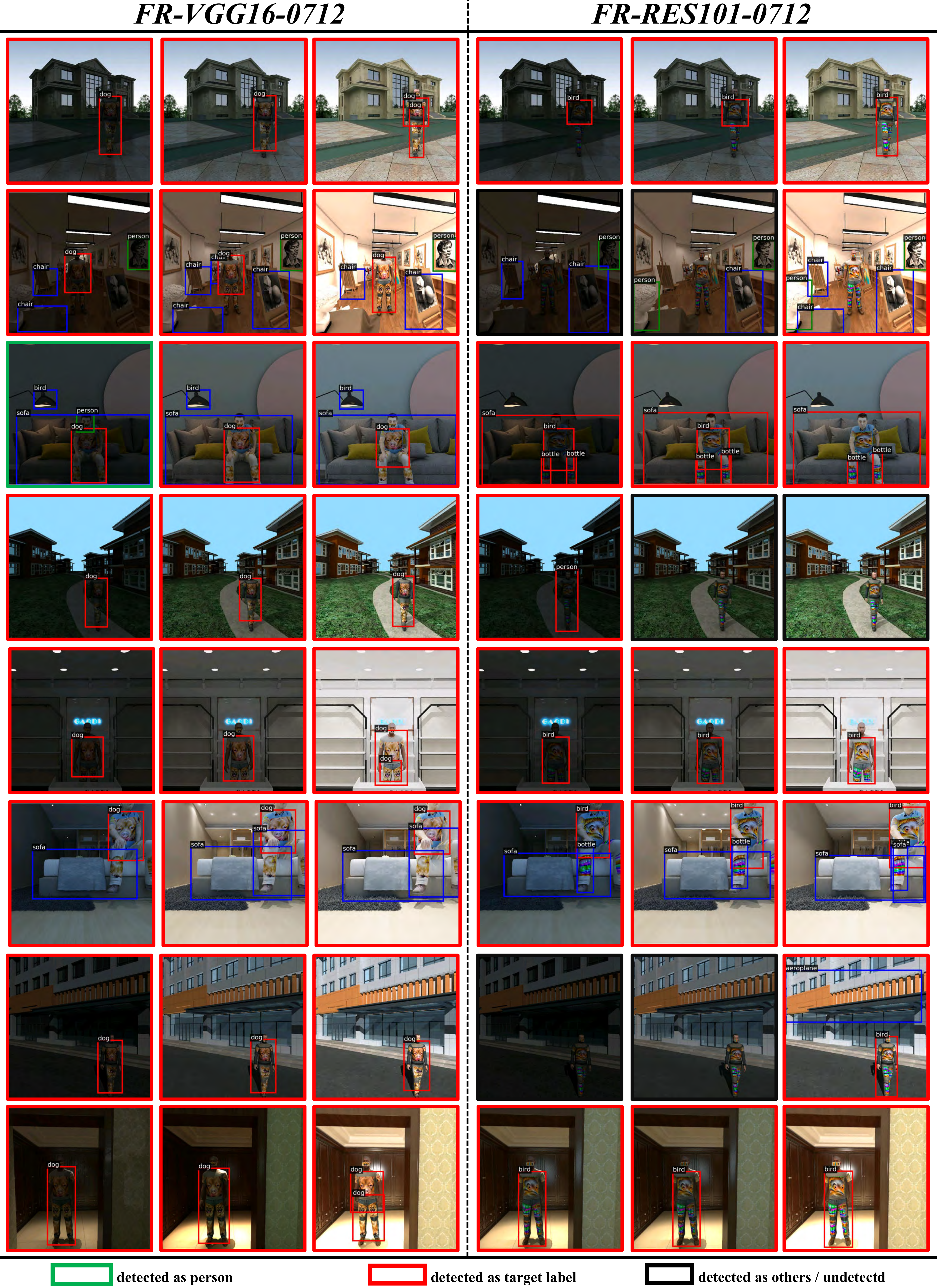}
	\end{center}
	\vspace{-0.5em}
	\caption{
		\textbf{More qualitative results of FR-VGG16-0712 and FR-RES101-0712 on virtual environment.} Each row set different virtual environments with the same viewpoint of camera, and each column uses different lighting condition.
	}
	\label{fig:virtual_scenes}
	\vspace{-0.5em}
\end{figure*}

Sampled results captured in different physical conditions (\eg, brightness, background environments) are shown in Figure \ref{fig:virtual_scenes}, which further show the power of proposed method.
\clearpage

\subsection{Experiment Results of Other Labels}
\begin{figure*}[h]
	\begin{center}
		\includegraphics[width=0.8\linewidth]{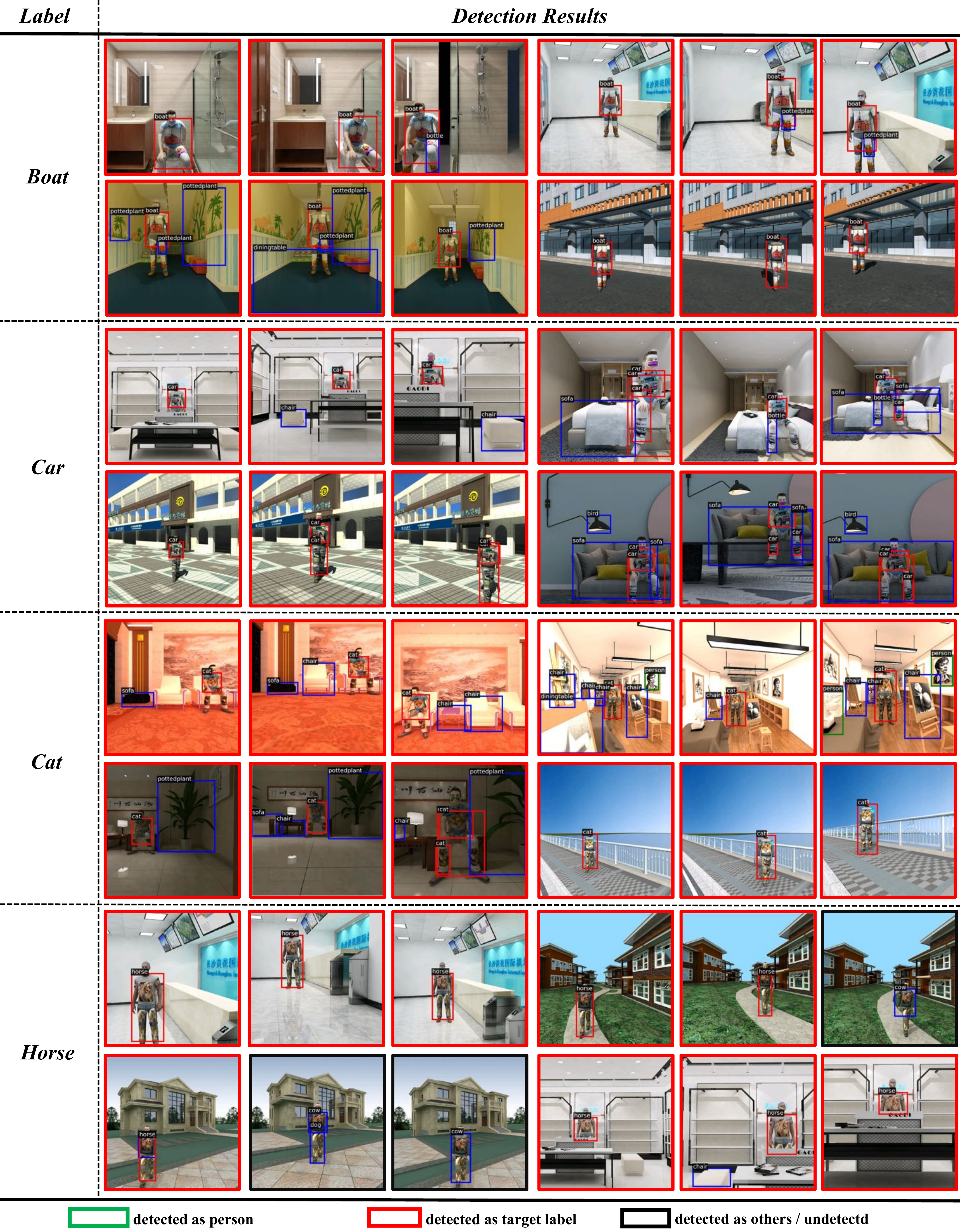}
	\end{center}
	\vspace{-0.5em}
	\caption{
		\textbf{More qualitative results of targeting other categories.} Each row applies different patterns (\ie, boat/car/cat/horse), and captured in different viewpoints and background environments.
	}
	\label{fig:other_labels}
	\vspace{-0.5em}
\end{figure*}

We show some results of targeting other categories in Figure~\ref{fig:other_labels}.
\clearpage

\section{Visualization Study}
\begin{figure*}[h]
	\begin{center}
		\includegraphics[width=0.8\linewidth]{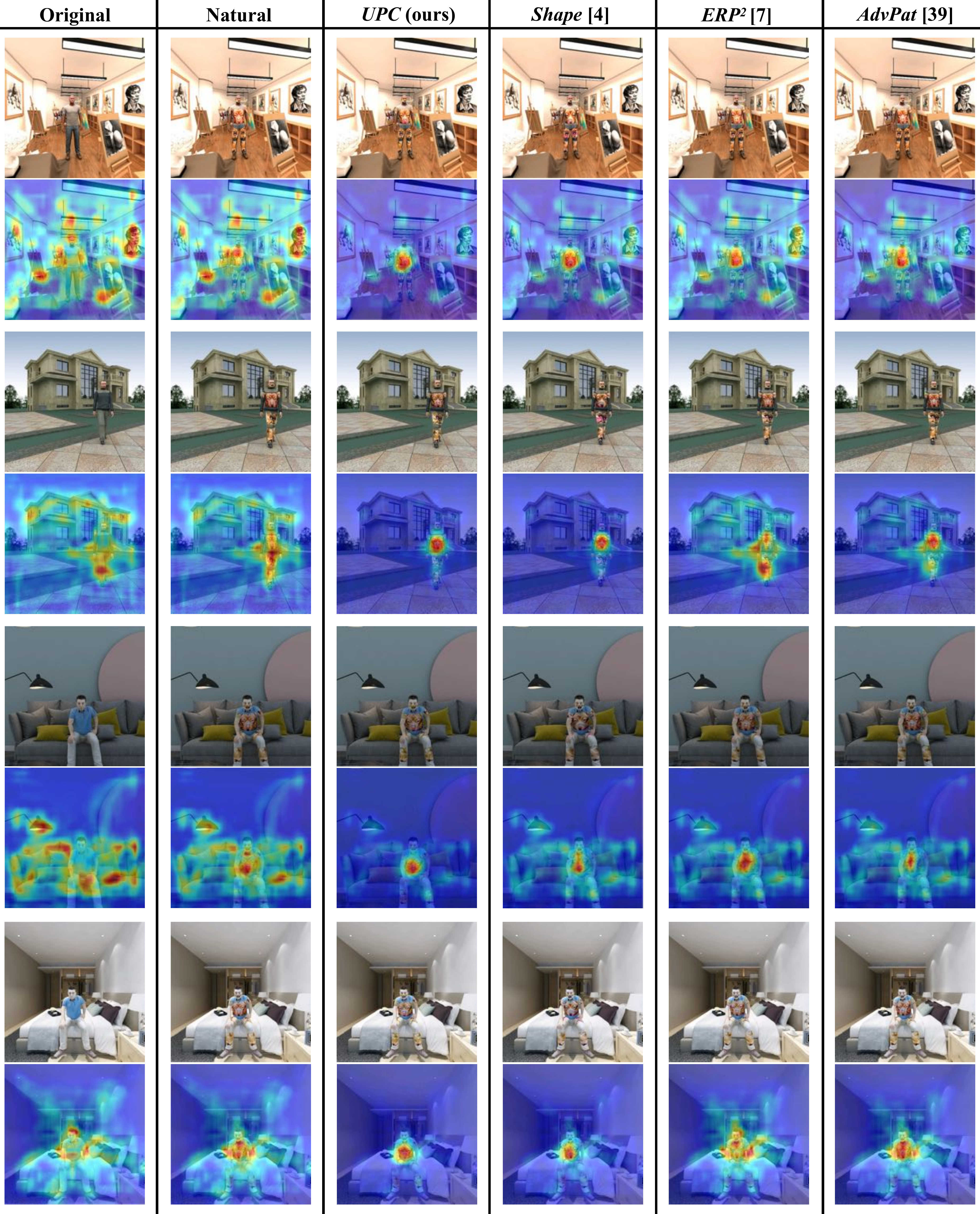}
	\end{center}
	\vspace{-0.5em}
	\caption{
		\textbf{Visualization of discriminative regions between different methods (\ie, $Shape$, $ERP^2$, $AdvPat$ and $UPC$).} The rendered images include Standing/Walking/Sitting poses, and captured in different viewpoints and background environments.
	}
	\label{fig:visualize_methods}
	\vspace{-0.5em}
\end{figure*}

In order to study the intrinsic reasons for different performance between existing methods, we visualize the feature maps from last convolutional layer to represent discriminative regions (see Figure~\ref{fig:visualize_methods}). We observe that natural scheme can not change the detectors attention though patterns occlude some parts of human body. On the contrary, total four attacks attempt to fool detectors by activating patterns' feature. However, the visualization result demonstrate existing methods (\ie, $Shape$, $ERP^2$, $AdvPat$) can not depress the activated features of un-occluded parts (\ie, face, hand) effectively, which may lead higher detection accuracy.

\clearpage

\begin{figure*}[h]
	\begin{center}
		\includegraphics[width=0.8\linewidth]{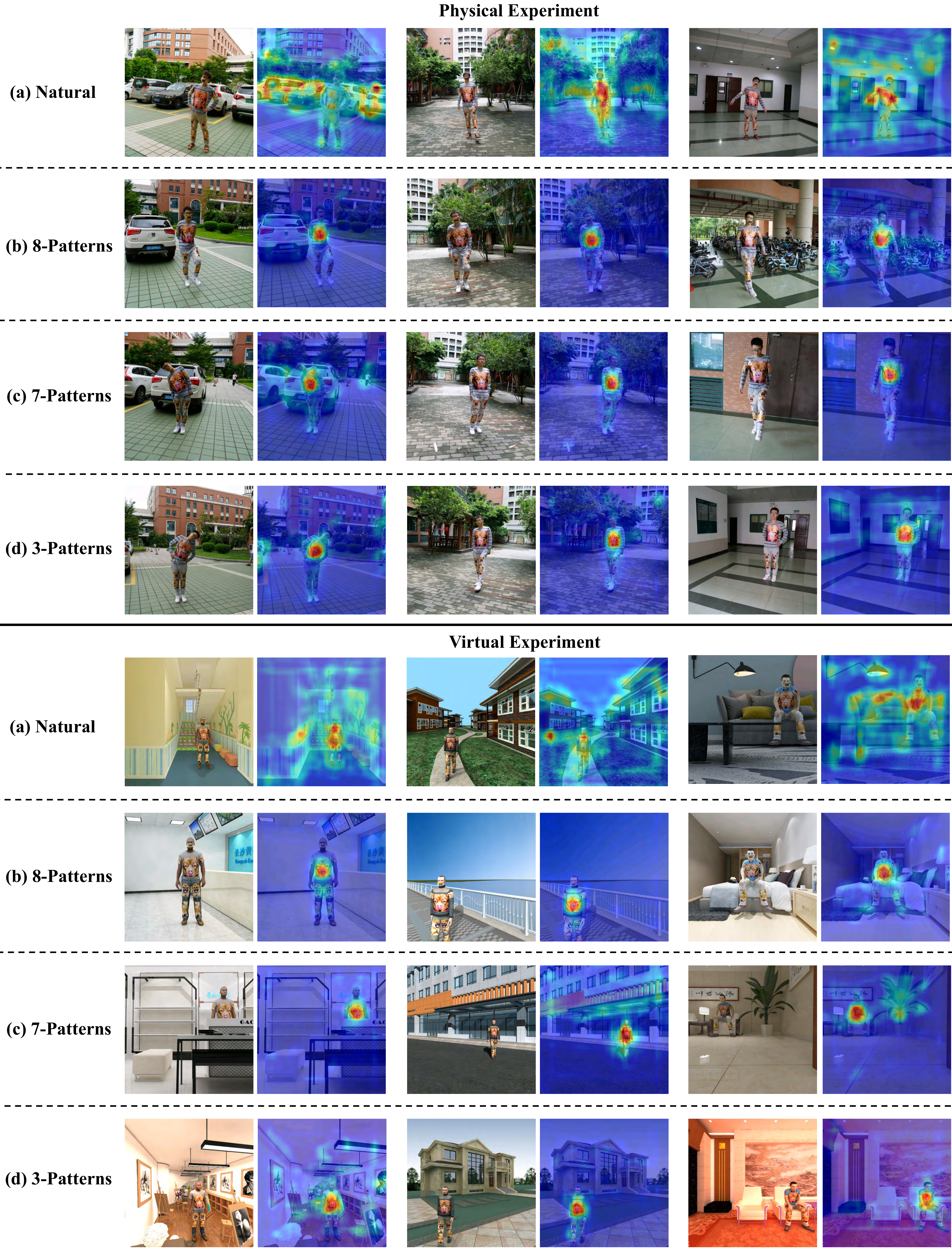}
	\end{center}
	\vspace{-0.5em}
	\caption{
		\textbf{Visualization of discriminative regions of proposedd UPC between different pattern schemes and poses.} Both physical and virtual results demonstrate similar trends under various physical conditions.
	}
	\label{fig:visualize_all}
	\vspace{-0.5em}
\end{figure*}

Here we show some visualization results for different pattern schemes and poses both in physical world and virtual scenes (Figure~\ref{fig:visualize_all}). We can observe that the feature of face or hands will be activated when less surface area covered. 

\clearpage
\section{Generalization to Other Categories}
In this section, we show some qualitative results of UPC on fooling the ``car'' category in both virtual scenes and physical world. Video of this experiment is available in the supplemental files.
\begin{figure*}[h]
	\begin{center}
		\includegraphics[width=0.65\linewidth]{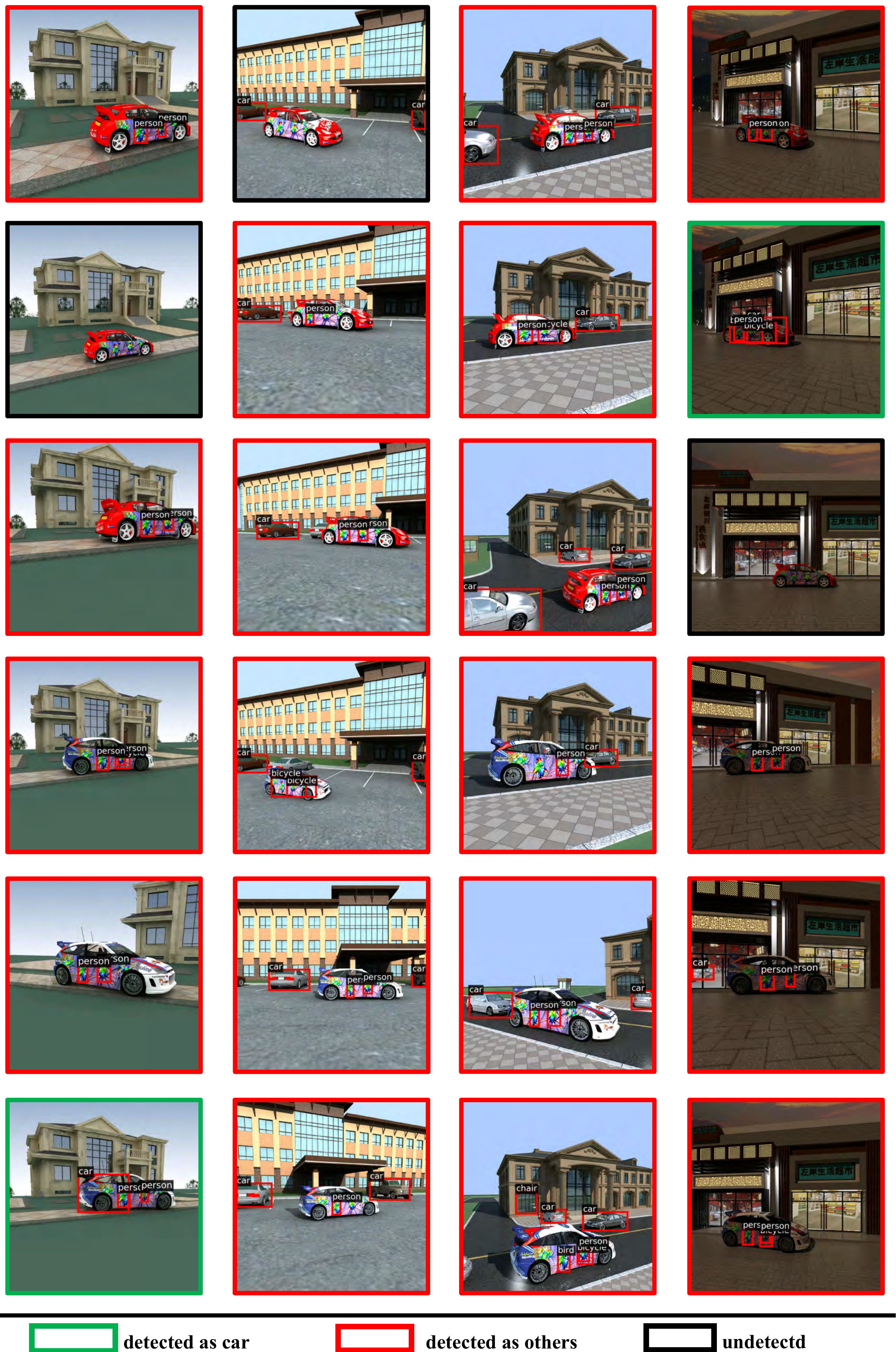}
	\end{center}
	\vspace{-0.5em}
	\caption{
		\textbf{More qualitative results of attacking the ``car'' category in virtual scenes.}
		We use two different car models (red car in top three rows and white car in bottom three rows) to evaluate the generalizablity of UPC.
	}
	\label{fig:car_compare}
	\vspace{1.em}
\end{figure*}

\begin{figure*}[h]
	\begin{center}
		\includegraphics[width=0.9\linewidth]{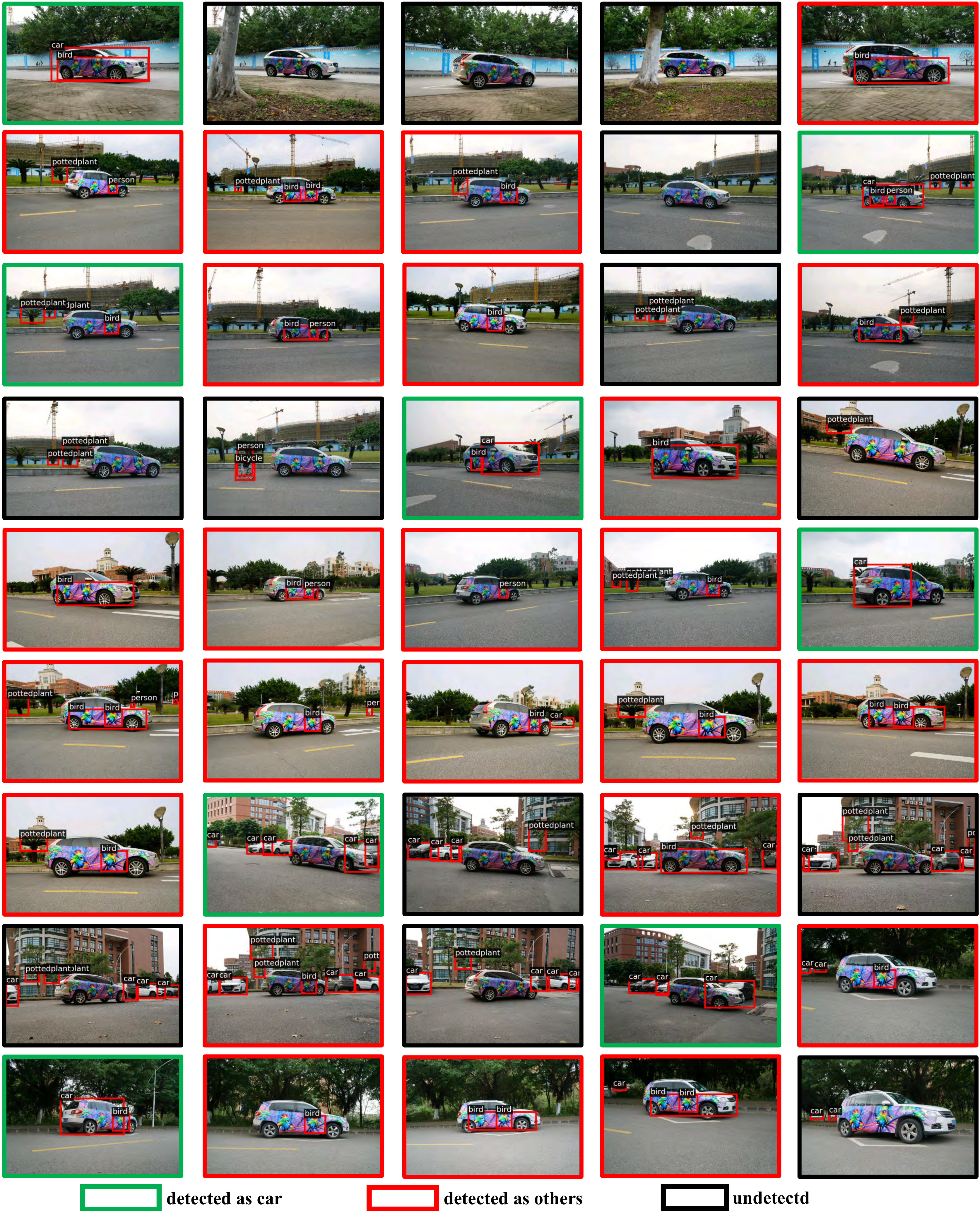}
	\end{center}
	\vspace{-0.5em}
	\caption{
		\textbf{More experimental results of fooling the ``car'' category in physical world.} We attack two different cars, \ie, Volvo XC60 and Volkswagen Tiguan.}
	\label{fig:physical_car}
	\vspace{-0.5em}
\end{figure*}